\documentclass[10pt,twocolumn,letterpaper]{article}
\usepackage{3dv} % DO NOT CHANGE THIS
\usepackage{times}  % DO NOT CHANGE THIS
\usepackage{epsfig}
\usepackage{graphicx}
\usepackage{amsmath,amssymb} % define this before the line numbering.
\usepackage{color}

\usepackage{algorithm,algorithmicx,algpseudocode}
\usepackage{bm,xspace}
\usepackage{comment}
\usepackage{verbatim}
\usepackage{balance}
\usepackage{booktabs}
\usepackage{etoolbox,siunitx}
\usepackage{calc}
\usepackage{pifont,hologo}
\usepackage{adjustbox}
\usepackage{tabularx}
\usepackage{multirow}
\usepackage{caption} % enable sub-table customized caption name
\usepackage{overpic}
\usepackage{array,makecell} 
\usepackage{esvect} % Extended Vector Arrow
\usepackage{gensymb} % for \degree 
\usepackage[table]{xcolor}

\definecolor{mygray}{gray}{0.9} % define your gray;
\definecolor{myskyblue}{RGB}{230, 249,255}
\definecolor{myyellowgreen}{RGB}{240, 240, 245}
\usepackage[pagebackref=true,breaklinks=true,colorlinks,bookmarks=false]{hyperref}

\threedvfinalcopy % *** Uncomment this line for the final submission

\def\threedvPaperID{108} % *** Enter the 3DV Paper ID here

\ifthreedvfinal\fi

\def\eg{\emph{e.g.~}}

\def\etal{\emph{et al.~}}

\newenvironment{packed_item}{
\begin{itemize}
\vspace{-0pt}
  \setlength{\itemsep}{0pt}
  \setlength{\parskip}{0pt}
  \setlength{\parsep}{0pt}
  \setlength{\topsep}{-10pt}
  \setlength{\partopsep}{0pt}
}{\end{itemize}}

\newcolumntype{?}[1]{!{\vrule width #1}} % thicker vertical line;
\newcolumntype{Y}{>{\centering\arraybackslash}X}
\newcommand*{\myqcrfont}{\fontfamily{<qcr>}\selectfont}
\title{Do End-to-end Stereo Algorithms Under-utilize Information?}
\author{
Changjiang Cai \\ {\tt\small ccai1@stevens.edu} 
\and 
Philippos Mordohai \\ {\tt\small mordohai@cs.stevens.edu }
}
\affiliation{Stevens Institute of Technology}
\begin{document}
%\nocopyright

\maketitle
%\thispagestyle{empty}

%%%%%%%%% ABSTRACT
\begin{abstract}

Deep networks for stereo matching typically leverage 2D or 3D convolutional encoder-decoder architectures to aggregate cost and regularize the cost volume for accurate disparity estimation. Due to content-insensitive convolutions and down-sampling and up-sampling operations, these cost aggregation mechanisms do not take full advantage of the information available in the images. Disparity maps suffer from over-smoothing near occlusion boundaries, and erroneous predictions in thin structures. In this paper, we show how deep adaptive filtering and differentiable semi-global aggregation can be integrated in existing 2D and 3D convolutional networks for end-to-end stereo matching, leading to improved accuracy. The improvements are due to utilizing RGB information from the images as a signal to dynamically guide the matching process, in addition to being the signal we attempt to match across the images. We show extensive experimental results on the KITTI 2015 and Virtual KITTI 2 datasets comparing four stereo networks (DispNetC, GCNet, PSMNet and GANet) after integrating four adaptive filters (segmentation-aware bilateral filtering, dynamic filtering networks, pixel adaptive convolution and semi-global aggregation) into their architectures. Our code is available at \url{https://github.com/ccj5351/DAFStereoNets}.
\end{abstract}

%-------------------------------------------------------------------------
%%%%%%%%% BODY TEXT
\section{Introduction}\label{sec:intro}

Progress in learning based stereo matching has been rapid from early work that focused on aspects of the stereo matching pipeline \cite{scharstein2002taxonomy}, such as similarity computation \cite{zbontar2016stereo}, to recent end-to-end systems \cite{chang2018psmnet,kendall2017-gcnet,mayer2016large,zhang2019ga}. 
What is remarkable is that the evolution of learning based stereo has largely mirrored that of conventional algorithms:
initial end-to-end systems resembled winner-take-all stereo with the disparity of each pixel determined almost independently based on a small amount of local context, while GA-Net \cite{zhang2019ga}, arguably the most effective current algorithm, includes a differentiable form of the Semi-Global Matching (SGM) algorithm \cite{hirschmuller08}, which has been by far the most popular conventional optimization technique for over a decade.
% PM: the above may need a little work to transition nicely to the next

Despite their success, %especially in settings where the training and test data do not differ substantially, 
deep stereo networks seem to under-utilize information present in their inputs. Specifically, in this paper we demonstrate how many existing networks leverage RGB information from the images to extract features that facilitate matching, but leave additional information unexploited. We show that the accuracy of representative network architectures can be improved by integrating into them modules that are sensitive to pixel similarity, image edges or semantics and act like adaptive filters.

In our experiments, we have integrated four components that can be thought of as adaptive or guided filters into four existing networks for stereo matching. Specifically, we have experimented with segmentation-aware bilateral filtering (SABF) \cite{harley_segaware2017}, dynamic filtering networks (DFN) \cite{jia2016dynamic}, pixel adaptive convolution (PAC) \cite{su2019pixel} and semi-global aggregation (SGA) from GANet \cite{zhang2019ga} integrated into DispNetC \cite{mayer2016large}, GCNet \cite{kendall2017-gcnet}, PSMNet \cite{chang2018psmnet}, and GANet \cite{zhang2019ga}. The set of filters is diverse: SABF, for example, is pre-trained to embed its input via a semantic segmentation loss, while SGA aggregates matching cost along rows and columns of cost volume slices. The backbone networks are also diverse, spanning 2D \cite{mayer2016large} and 3D \cite{chang2018psmnet,kendall2017-gcnet} convolutional networks and GANet that foregoes 3D convolutions in favor of the SGM-like aggregation mechanism. The number of parameters of the backbones ranges from 2.8 to 42.2 million. 

The contributions of this paper are:
\vspace{-8pt}
\begin{packed_item}
  \item Several novel deep architectures for stereo matching.
  \item Evidence that further progress in stereo matching is possible by leveraging image context as guidance for refinement, filtering and aggregation of the matching volume.
  \item A comparison of four filtering methods leading to the conclusion that SGA typically achieves the highest accuracy among them.
\end{packed_item}

%Our contributions can be summarized as follows: (1) Several novel deep architectures. (2) Finding that further progress in stereo matching by leveraging image context as guidance for refinement, filtering and aggregation of the matching volume. (3) A comparison of four filtering methods leading to the conclusion that SGA achieves the highest accuracy among them.

\section{Related Work}\label{sec:related}
%CCJ: make it shorter 
%In this section, we review two relevant threads of works: deep learning based end-to-end stereo matching and content guided and adaptive filtering techniques in CNNs. 

We review deep learning based end-to-end stereo matching and content-guided or adaptive filtering in CNNs. 

\vspace{3pt}\noindent \textbf{End-to-end Stereo Matching.} \, End-to-end stereo matching methods can be generally grouped into two categories: 
 2D CNNs for correlation-based disparity estimation and 3D CNNs for cost volume based disparity regression.
%i) 2D CNNs for correlation-based 3D cost volume;  ii) 3D CNNs for concatenation-based 4D cost volume.

%The first category starts with MC-CNN \cite{zbontar2016stereo} by training a CNN to learn the matching cost on small image patches, followed by several post-processing steps for disparity estimation, \eg CBCA \cite{zhang2009cross}, SGM \cite{hirschmuller08} etc. Several follow-up works developed MC-CNN. Using local expansion moves based on graph cuts, Taniai \cite{taniai2018local-expansion} \etal leveraged the matching cost to get SOTA result on Middlebury benchmark \cite{scharstein2014high}. Luo \etal \cite{luo2016efficient} replaced fully connected layers in MC-CNN with inner product layers for efficient computation.

A representative work in the first category is DispNetC \cite{mayer2016large}. %the first end-to-end stereo network. 
It computes the correlations among features between stereo views at different disparity values, and regresses the disparity via an encoder-decoder 2D CNN architecture. Many other methods \cite{liang2018learning_1st_Rob, pang2017cascade,  song2019edgestereo,Tonioni_2019_CVPR_madnet,yang2018segstereo,Yin_2019_CVPR_hd3,zhan2019dsnet} extend this paradigm. %by introducing new modules. 
CRL \cite{pang2017cascade} employs a two-stage network for cascade disparity residual learning. Feature constancy is added in iResNet \cite{liang2018learning_1st_Rob} to further refine the disparity. SegStereo \cite{yang2018segstereo}, DSNet \cite{zhan2019dsnet} and EdgeStereo \cite{song2019edgestereo} are multi-task learning approaches jointly optimizing stereo matching with semantic segmentation or edge detection. 

%SegStereo \cite{yang2018segstereo} exploiting semantic cues and EdgeStereo \cite{song2019edgestereo} incorporating edge cues help mitigate the disparity ambiguity in texture-less regions. [??? ``discuss receptive field here as a way to increase the context'' ???]

State-of-the-art stereo networks \cite{ chang2018psmnet, guo2019group, kendall2017-gcnet,zhang2019ga}  largely fall in the second category. Different than the correlation based cost volume in 2D-CNN stereo networks, 3D-CNNs generate a 4D cost volume by concatenating the deep features from the Siamese branches along the channel dimension at each disparity level and each pixel position. GCNet \cite{kendall2017-gcnet} and PSMNet \cite{chang2018psmnet} apply 3D convolutional layers for cost aggregation, followed by a differentiable \textit{soft-argmin} layer for disparity regression. GwcNet \cite{guo2019group} leverages group-wise correlation of channel-split features to generate a hybrid cost volume that can be processed by a smaller 
%for reduced amount of parameters in 
3D convolutional aggregation sub-network. 
GANet \cite{zhang2019ga} includes local guided aggregation (LGA) and semi-global aggregation (SGA) layers
for efficient cost aggregation which are complementary to 3D convolutional layers.
LGA aggregates the cost volume locally to refine thin structures, %and run at real-time speed, 
while SGA is a differentiable counterpart of SGM \cite{hirschmuller08}. 
%LGA runs at real-time speed to aggregate the cost volume in a local region to refine thin structures, while SGA foregoes 3D convolutions in favor of the SGM-like \cite{hirschmuller08} aggregation mechanism. SGA is detailed in Sec. \ref{subsec:sga}.

\vspace{3pt}
\noindent \textbf{Content Guided and Adaptive Filtering in CNNs.}  Existing content-adaptive CNNs fall into two general classes. In the first class, conventional image-adaptive filters (\eg bilateral filters \cite{aurich1995non, tomasi1998bilateral}, guided image filters \cite{he2012guided} and non-local means \cite{awate2005higher, buades2005non}, among others) have been adapted to be differentiable and used as content-adaptive neural network layers \cite{chandra2016_eccv_fast,chen2016_cvpr_semantic, chen2015_ICML_learning, gadde16bilateralinception, jampani2016_cvpr_learningSparse, krahenbuhl2011efficient,lin2016_cvpr_efficient,liu2017nips_learning_affinity, wang2018_cvpr_non-local, wu2018cvpr_fast_guided_filter, zheng_crfasrnn_ICCV2015}. Wu \etal \cite{wu2018cvpr_fast_guided_filter} propose novel layers to perform guided filtering \cite{he2012guided} inside CNNs. Wang \etal \cite{wang2018_cvpr_non-local} present non-local neural networks to mimic non-local means \cite{buades2005non} for capturing long-range dependencies. The bilateral inception module by Gadde \etal \cite{gadde16bilateralinception} can be inserted into existing CNN segmentation architectures for improved results. It performs bilateral filtering to propagate information between superpixels based on the spatial and color similarity. Harley \etal \cite{harley_segaware2017} integrate segmentation information in the CNN by firstly learning segmentation-aware embeddings, then generating local foreground attention masks, and combining the masking filters with convolutional filters to perform segmentation-aware convolution (see details in Sec. \ref{subsec:ebf}).  Deformable convolutions 
% PM commented Dai_2017_ICCV_deformable_cnn out
% CCJ: add it back and comment it out in other place.
\cite{ Dai_2017_ICCV_deformable_cnn, Zhu_2019_CVPR} produce spatially varying modifications to the convolutional filters, where the modifications are represented as offsets in favor of learning geometric-invariant features. Pixel adaptive convolution (PAC) \cite{su2019pixel} mitigates the content-agnostic drawback of standard convolutions by multiplying the convolutional filter weights by a spatial kernel function. %The kernel has a predefined form (\eg Gaussian) and guarantee to response differently depending on the position and features of local pixels. 
PAC has been applied in joint image upsampling networks \cite{DJF-ECCV-2016} and in a learnable dense conditional random field (CRF) framework \cite{chen2015_ICML_learning,jampani2016_cvpr_learningSparse,krahenbuhl2011efficient,zheng_crfasrnn_ICCV2015}. See Sec. \ref{subsec:pac} for more details.

%Conditional random field (CRF) approach has been integrated with the CNN, by framing the CRF as a recurrent network, and chaining it to the backpropagation of the underlying CNN \cite{zheng_crfasrnn_ICCV2015}. 

%The aforementioned SGA \cite{zhang2019ga} makes SGM \cite{hirschmuller08} differentiable so as to filter cost volume in end-to-end stereo networks. 

Another class of content-adaptive CNNs 
focuses on learning spatial position-aware filter weights using separate sub-networks. These approaches are called ``Dynamic Filter Networks'' (DFN) 
% CCJ: commented Dai_2017_ICCV_deformable_cnn out
\cite{jia2016dynamic,Wu_2018_ECCV_dynamicFilter,xue2016_nips_visual} or kernel prediction networks \cite{bako2017kernel}, which have been used in several computer vision tasks.
%, \eg video and stereo prediction.  -- PM: what does this even mean?
Jia \etal \cite{jia2016dynamic} propose the first DFN where the convolutional filters are generated dynamically depending on input pixels. The filter weights, provided by a filter-generating network conditioned on an input, are applied to another input through the dynamic filtering layer (see details in Sec. \ref{subsec:dfn}). It is extended by Wu \etal \cite{Wu_2018_ECCV_dynamicFilter} with an additional attention mechanism and a dynamic sampling strategy to allow the position-specific kernels to also learn from multiple neighboring regions.
% Similar ideas have been applied to more specific tasks, \eg motion prediction \cite{xue2016_nips_visual}, semantic segmentation \cite{harley_segaware2017}, and Monte Carlo rendering denoising \cite{bako2017kernel}.

% Above is fine
% Y. Li, J.-B. Huang, N. Ahuja, and M.-H. Yang. Deep Joint Image Filtering. In ECCV, 2016
% J.T. Barron and B. Poole. The Fast Bilateral Solver. In ECCV, 2016.
% K. He, J. Sun, and X. Tang. Guided Image Filtering. PAMI, 35:1397–1409, 06 2013 --- inspired by

\section{Approach}\label{sec:approach}
%Comments: It should have a section on the embedding and filtering methods and a second section on the architectures. The architectures should be described in less detail, focusing on how we integrated the filters/embedding in the original networks.

In this section, we describe our approach for adapting state-of-the-art stereo matching networks \cite{chang2018psmnet,kendall2017-gcnet,  mayer2016large, zhang2019ga} by integrating deep filtering techniques  %\cite{harley_segaware2017,jia2016dynamic,su2019pixel, zhang2019ga} 
to improve their accuracy. We show how four filtering techniques, segmentation-aware bilateral filtering (SABF) \cite{harley_segaware2017}, dynamic filtering networks (DFN) \cite{jia2016dynamic}, pixel adaptive convolution (PAC) \cite{su2019pixel} and semi-global aggregation (SGA) \cite{zhang2019ga},
%dynamic filtering layer (DFL) in dynamic filter network (DFN) \cite{jia2016dynamic}, pixel adaptive convolution (PAC) operation \cite{su2019pixel} and semi-global aggregation (SGA) from GA-Net \cite{zhang2019ga}, 
can be used to filter the cost volume for accurate disparity estimation.

\subsection{Cost Volume in Stereo Matching}\label{sec:cv}

\begin{figure*}
\begin{center}
\includegraphics[width=0.9\linewidth]{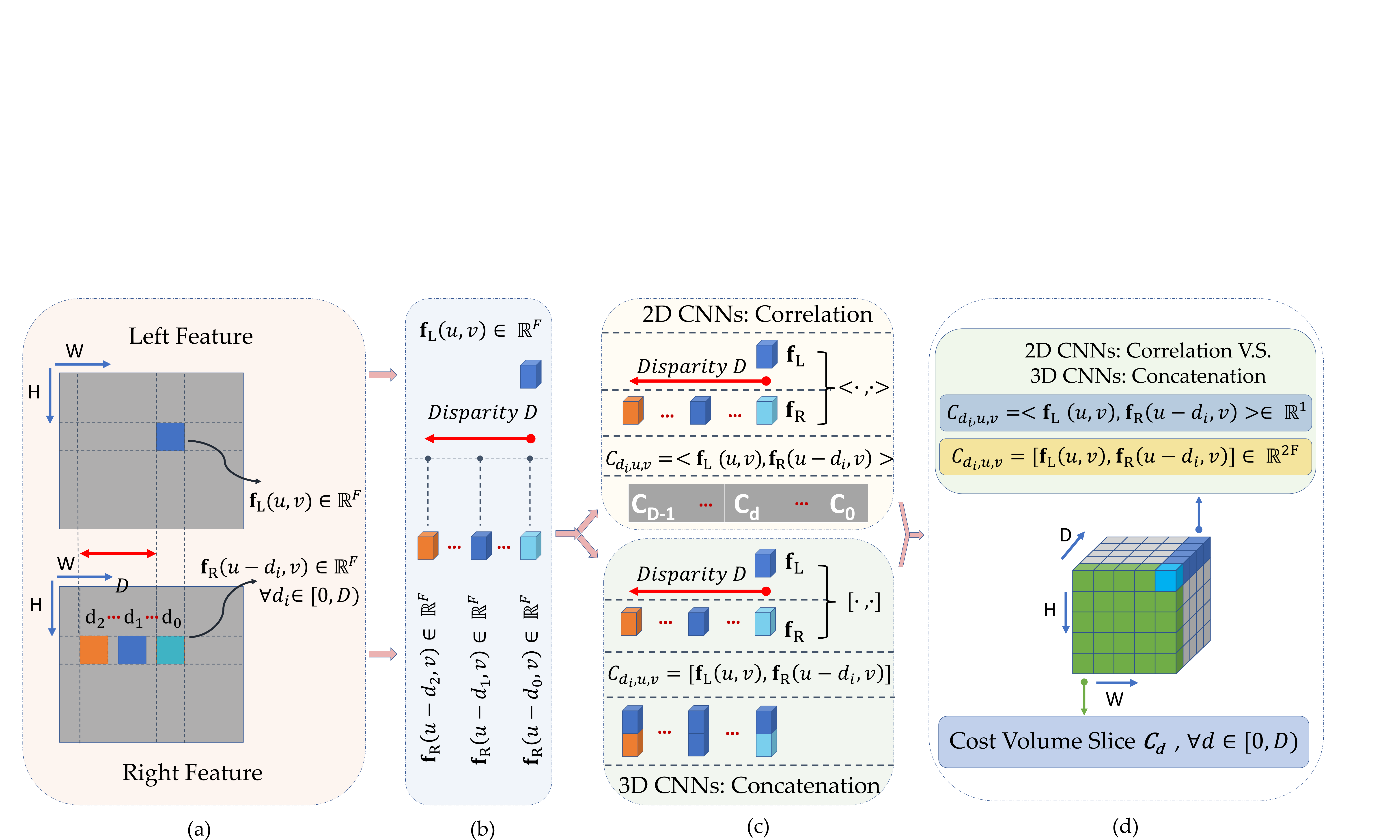}
\end{center}
\vspace{-12pt}
\caption{ 
Cost volume in 2D and 3D CNNs for stereo matching. (a) Deep features are extracted for each view. (b) A left feature vector $\mathbf{f}_L(u,v)$ and several counterparts $\mathbf{f}_R(u-d_i,v)$, at disparity $d_i \in [0, D)$. (c) The left feature vector is either correlated (top branch for 2D CNNs) or concatenated (bottom branch for 3D CNNs) with the corresponding right feature vectors. (d) The cost volume, including the highlighted cost volume slice $\mathcal{C}_d$ (green), cost volume fiber $\mathcal{C}_{u,v}$ (blue) and cost feature $\mathcal{C}_{d, u, v}$ (light blue) that is a scalar for 2D CNNs or a vector for 3D CNNs. 
}
\label{fig:cost-volume-generation}
\end{figure*}

Given a rectified stereo pair $\mathbf{I}_L$ and $\mathbf{I}_R$ with dimensions $H \times W$ ($H$: height, $W$: width), stereo matching finds the correspondence between a reference pixel $\mathbf{p}_L$ in the left image $\mathbf{I}_L$ and a target pixel $\mathbf{p}_R$ in the right image $\mathbf{I}_R$. The cost volume (or matching volume) $\mathcal{C}$ is defined as a 3D or 4D tensor with dimensions $D \times H \times W$ or $2F \times D \times H \times W$ ($D$: disparity range, $F$: feature dimensionality for each of the views), to represent the likelihood of a reference pixel $\mathbf{p}_L(u,v)$ corresponding to a target pixel $\mathbf{p}_R(u-d,v)$, with disparity $d \in [0, D)$.
%Given a rectified input stereo pair $\mathbf{I}_L$ and $\mathbf{I}_R$ with dimensions $H \times W$ ($H$: height, $W$: width) in stereo matching, the cost volume (or matching volume) $\mathcal{C}$ is defined as a 3D or 4D tensor with dimensions $D \times H \times W$ or $2F \times D \times H \times W$ ($D$: disparity range, $F$: feature dimensionality for each of the stereo views), to represent the likelihood of a pixel $P_L(x,y)$ in the reference image (which is typically the left image) $\mathbf{I}_L$ corresponding to a pixel $P_R(x-d,y)$ (where $d \in [0, D)$) in the target image $\mathbf{I}_R$. 
Its construction is illustrated in Fig \ref{fig:cost-volume-generation}. Specifically, the extracted left feature vector %$\vv{f_L}(x,y) \in \mathcal{R}^F$
$\mathbf{f}_L(u,v) \in \mathbb{R}^F$ of the reference pixel $\mathbf{p}_L(u,v)$ is either \textbf{correlated} (in 2D CNNs \eg DispNetC \cite{mayer2016large} and iResNet \cite{liang2018learning_1st_Rob}) or \textbf{concatenated} (in 3D CNNs \eg GCNet \cite{kendall2017-gcnet}, PSMNet \cite{chang2018psmnet} and GANet \cite{zhang2019ga}) with the corresponding feature $\mathbf{f}_R(u-d,v) \in \mathbb{R}^F$
of the target pixel $\mathbf{p}_R(u-d,v)$, for each disparity $d \in \{ 0, \dots, D-1 \}$.
%$d \in [0, D)$. -- this notation implies an interval of real numbers. replace with d \in {0, ..., D-1}
It is formulated in Eq.  %\ref{eq:cv-2d-generation} and \ref{eq:cv-3d-generation}, respectively.
\ref{eq:cv-generation}:

%\begin{equation}\label{eq:cv-2d-generation}
%C_{d, x, y} =  \vv{f_L}(x,y) \bigodot \vv{f_R}(x-d,y) \in \mathcal{R}^1 
%\end{equation}

%\begin{equation}\label{eq:cv-3d-generation}
%C_{d, x, y} =  \vv{f_L}(x,y) || \vv{f_R}(x-d,y)  \in \mathcal{R}^{2F}
%\end{equation}

\vspace{-12pt}
\begin{equation}\label{eq:cv-generation}
\mathcal{C}_{d, u, v} =  
\begin{cases} 
      %\vv{f_L}(x,y) \bigodot \vv{f_R}(x-d,y) \in \mathcal{R}^1 &\text{2D CNNs} \\
      % \ast or \star
      %\mathbf{f}_L(x,y) \ast \mathbf{f}_R(x-d,y) \in \mathbb{R}^1 &\text{2D CNNs} \\
      \langle \mathbf{f}_L(u,v), \mathbf{f}_R(u-d,y) \rangle \in \mathbb{R}^1 &\text{\small{2D CNNs}} \\
      %\mathbf{f}_L(x,y) || \mathbf{f}_R(x-d,y)  \in \mathbb{R}^{2F} &\text{3D CNNs}
      [\mathbf{f}_L(u,v), \mathbf{f}_R(u-d,v)]  \in \mathbb{R}^{2F} &\text{\small{3D CNNs}}\\
   \end{cases}
\end{equation}

\noindent where $\langle \cdot, \cdot \rangle$ denotes correlation, $[\cdot, \cdot]$ indicates concatenation, 
%disparity $d \in [0, D)$, image coordinates $u \in [0, W)$ and $v \in [0, H)$, PM: repetition inconsistent with two lines below
resulting in a 3D or 4D tensor for 2D and 3D CNNs, respectively. We denote a \textbf{cost volume slice} at disparity $d$ by  $\mathcal{C}_d$ (i.e., $\mathcal{C}_d = \mathcal{C}_{d, u, v} |_{ u=1: W; v=1:H}$).

\subsection{Content-Adaptive Filtering Modules}
In this subsection, we present four content-adaptive filtering approaches which can be %simply yet 
effectively incorporated as content-adaptive CNN layers in state-of-the-art networks for end-to-end stereo matching.

%%%%%%%%%%%%%%%%%%%%%%%%%%%%%%%%%%%%%%%%%%
%%%%%%%%%%%%%%%%%  SABF  %%%%%%%%%%%%%%%%%
%%%%%%%%%%%%%%%%%%%%%%%%%%%%%%%%%%%%%%%%%%
\subsubsection{3.2.1 Segmentation-aware Bilateral Filtering Module} \label{subsec:ebf}
\begin{figure}
\begin{center}
\includegraphics[width=0.9\linewidth]{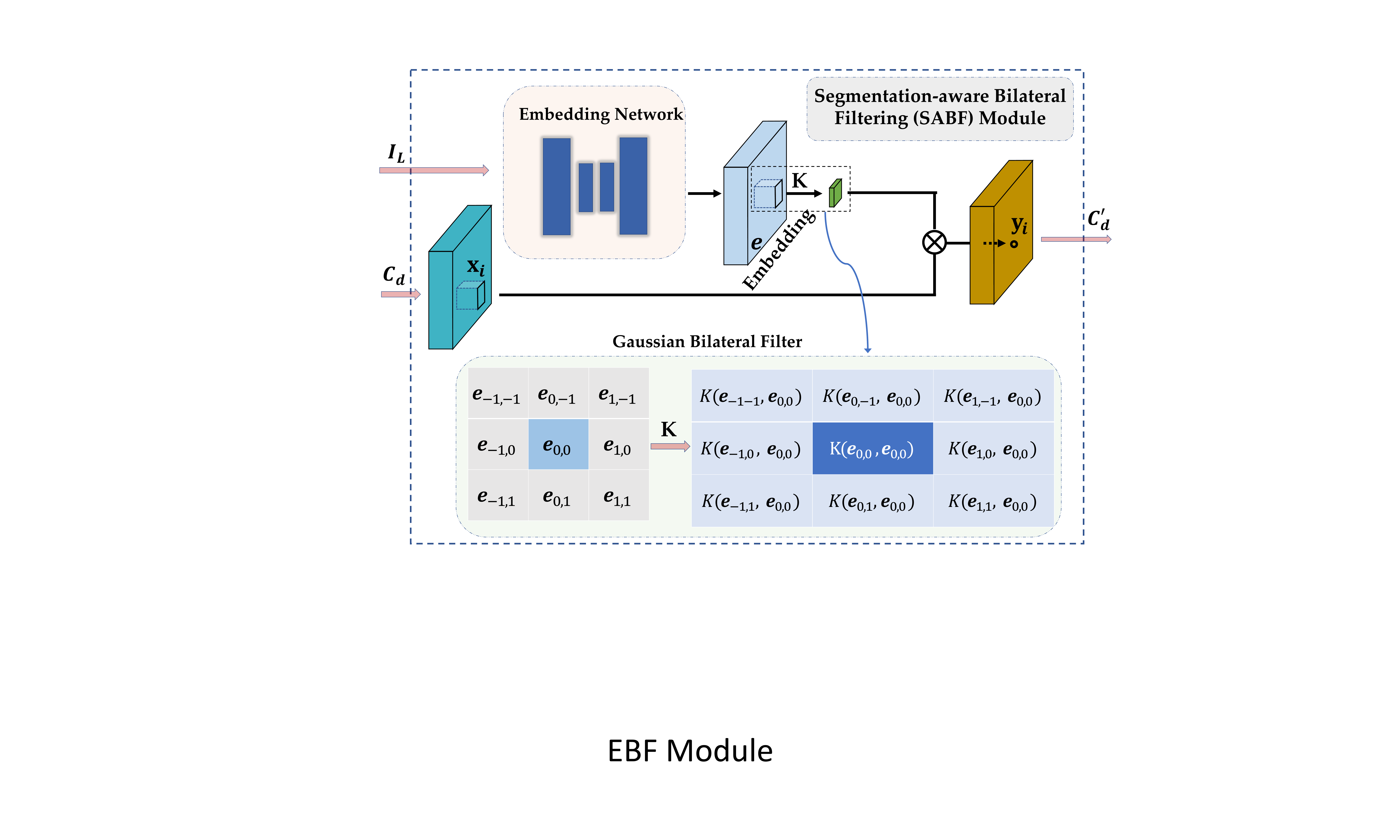}
\end{center}
\vspace{-9pt}
\caption{ 
Integrating the segmentation-aware bilateral filtering (SABF) module.  The upper branch shows the embedding $\mathbf{e}$ is learned by an \textit{embedding network} from an input image $\mathbf{I}_L$. Pairwise embedding distances are converted (Eq. \ref{eq:ebf-filter}) to SABF filter weights $K$, as shown at the bottom. The %middle branch 
overall figure
shows how SABF %(Eq. \ref{eq:ebf-layer})  
filters a cost volume slice $\mathcal{C}_d$ to obtain a segmentation-aware filtered result $\mathcal{C}^{\prime}_d$.}
\label{fig:sabf}
\end{figure}

Segmentation-aware bilateral filtering (SABF) \cite{harley_segaware2017} was proposed to enforce smoothness while preserving region boundaries or motion discontinuities in dense prediction tasks, such as semantic segmentation and optical flow estimation. As shown in Fig. \ref{fig:sabf}, here we adapt the SABF to stereo matching to filter the cost volume $\mathcal{C}$ (Sec. \ref{sec:cv}), by (i) learning to embed in a feature space where semantic dissimilarity between pixels can be measured by a distance function \cite{chopra2005learning}; (ii) creating local foreground (relative to a given pixel) attention filters $K^{sabf}$; (iii) filtering the cost volume $\mathcal{C}$ so as to capture the relevant foreground and be robust to appearance variations in the background or occlusions.

%\paragraph{Embedding Learning.}
\vspace{3pt}\noindent \textbf{Learning the Embedding.}  \, Given an RGB image $\mathbf{I}$, consisting of $N$ pixels $\mathbf{q}$, 
and its semantic segmentation label map $\mathbf{l}$, the embedding function is implemented as an \textit{Embedding Network} (see the detailed architecture in supplement) that maps pixels into an embedding space as $f : \mathbb{R}^3 \mapsto \mathbb{R}^{64}$, or $f(\mathbf{q}) = \mathbf{e}$, where $\mathbf{e} \in \mathbb{R}^{64}$ is the embedding for RGB pixel $\mathbf{q}$, with $64$ as the dimension of the embedding space. See Fig. \ref{fig:sabf}.
The embeddings are learned via a loss function over pixel pairs sampled in a neighborhood around each pixel. Specifically, for any two pixels $\mathbf{p}_i$ and $\mathbf{p}_j$ 
and corresponding object class labels % CCJ: added for I_i, I_j
$\mathbf{l}_i$ and $\mathbf{l}_j$, the pairwise loss is:
%$\mathbf{p}_i = (u_i, v_i)$ and $\mathbf{p}_j = (u_j, v_j)$, the pairwise loss is defined as in Eq. \ref{eq:embedding}: -- PM coordinates never appear

%we can optimize the same-label pairs to have “near” embeddings, and the different-label pairs to have “far” embeddings. Using $\alpha$ and $\beta$ to denote the “near” and “far” thresholds, respectively, we can define the pairwise loss as in Eq. \ref{eq:embedding} ... 

\vspace{-12pt}
\begin{equation}\label{eq:embedding}
\ell_{i,j} = 
\begin{cases} 
 \text{ } \max (|| \mathbf{e}_i - \mathbf{e}_j ||_1 - \alpha, 0) & \text{if } \mathbf{l}_i = \mathbf{l}_j \\
 \text{ } \max (\beta - ||\mathbf{e}_i - \mathbf{e}_j||_1, 0) & \text{if } \mathbf{l}_i \neq  \mathbf{l}_j \\
\end{cases}
\end{equation}

\noindent where $|| \cdot||_1$ indicates the $L_1$-norm, and the thresholds are $\alpha = 0.5$ and $\beta = 2$. Therefore, the total loss for all the pixels in the image $\mathbf{I}$ is defined in Eq. \ref{eq:embeddingLoss}:

\begin{equation}\label{eq:embeddingLoss}
\mathcal{L} = \sum_{i=0}^{N-1}\sum_{j \in \mathcal{N}_i} \ell_{i,j} 
\end{equation}

\noindent where $j \in \mathcal{N}_i$ spans the spatial neighbors of index $i$. We follow the implementation of Harley \etal \cite{harley_segaware2017} where three overlapping $3 \times 3$ neighborhoods with dilation rates of $1$, $2$, and $5$ are used for a good trade-off between long-range pairwise connectivity and computational efficiency. 
%for a good trade-off between receptive field's size and computational efficiency. 

\vspace{3pt}\noindent \textbf{Applying the SABF Layer.}  \, Once the embedding is learned, the SABF filter weights are obtained by converting the pairwise distance between $\mathbf{e}_i$ and $\mathbf{e}_j$ into (unnormalized) probabilities using two Gaussian distributions as in Eq. \ref{eq:ebf-filter}:

\begin{equation}\label{eq:ebf-filter}
K^{sabf}_{i,j}  = \exp{\left(  - \frac{||\mathbf{p}_i- \mathbf{p}_j ||^2}{2\sigma_s^2}  - \frac{||\mathbf{e}_i - \mathbf{e}_j||^2}{2\sigma_r^2} \right)}
\end{equation}

\noindent where $\sigma_s$ and $\sigma_r$ are two predefined standard deviations. Then, given an input feature $\mathbf{x}_i$, we can efficiently compute a segmentation-aware filtered result $\mathbf{y}_i$ via the SABF layer:
%implemented as in Eq. \ref{eq:ebf-layer}:

%\begin{equation}\label{eq:ebf-layer}
%\mathbf{y}_i = \frac{\sum_{k \in \Omega(i)} \mathbf{x}_{i-k} K^{ebf}_{i,i-k}}{\sum_{k \in %\Omega(i)}K^{ebf}_{i,i-k} }
%\end{equation}
\begin{equation}\label{eq:ebf-layer}
\mathbf{y}_i = \frac{\sum_{k \in \Omega(i)} \mathbf{x}_{k} \, K^{sabf}_{i,k}}{\sum_{k \in \Omega(i)}K^{sabf}_{i,k} }
\end{equation}

\noindent where $\Omega(\cdot)$ defines an $s \times s$ (e.g., $5 \times 5$) filtering window. The input $\mathbf{x}_i$ and output $\mathbf{y}_i$ here are from a raw cost volume slice $\mathcal{C}_d$ and the filtered slice
%$\widetilde{\mathcal{C}}_d$
$\mathcal{C}^{\prime}_d$, respectively (Fig. \ref{fig:sabf}).

%%%%%%%%%%%%%%%%%%%%%%%%%%%%%%%%%%%%%%%%%%
%%%%%%%%%%%%%%%%%  DFN   %%%%%%%%%%%%%%%%%
%%%%%%%%%%%%%%%%%%%%%%%%%%%%%%%%%%%%%%%%%%
\subsubsection{3.2.2 Dynamic Filtering Networks (DFN) Module} \label{subsec:dfn}
Dynamic Filter Networks (DFN) %are proposed by Jia \etal 
\cite{jia2016dynamic} are a content-adaptive filtering technique. As shown in Fig. \ref{fig:our-net}(b), the filters $\mathcal {F}_\theta$ in the DFN are dynamically generated by a separate \textit{filter-generating network} conditioned on an input $\textbf{x}_A$. Then, they are applied to another input $\mathbf{x}_B$ via the \textit{dynamic filtering layer}. In our implementation, $\mathbf{x}_A$ is the deep feature $\mathbf{f}_L$ of the reference image $\mathbf{I}_L$, and $\mathbf{x}_B$ is a cost volume slice $\mathcal{C}_d$.

\vspace{3pt} \noindent  \textbf{Filter-Generating Network.} \, Given an input $\mathbf{x}_A \in \mathbb{R}^{H \times W  \times C_A}$ ($H$: height, $W$: width, $C_A$: channel size of $\mathbf{x}_A$), the filter-generating network generates dynamic filters $\mathcal {F}_\theta$, parameterized by $\theta \in \mathbb{R}^{s \times s  \times C_B  \times n_{F}}$ ($s$: filter window size, $C_B$: channel size of $\mathbf{x}_B$, $n_{F}$: the number of filters). 
% PM not here
%The filter window size $s$ decides the receptive field. In our experiments, we find $s=5$ (with a dilation factor of $2$) works well enough. 
%The size of the receptive field can also be increased by stacking multiple dynamic filter modules. This is for example useful in applications that may involve large local displacements.

\vspace{3pt} \noindent \textbf{Dynamic Local Filtering Layer.} \, The generated filters $\mathcal {F}_\theta$ are applied to input images or feature maps $\mathbf{x}_B \in \mathbb{R}^{H \times W \times C_B}$ via the \textit{dynamic filtering layer} to output the filtered result $G = \mathcal {F}_\theta (\mathbf{x}_B) \in \mathbb{R}^{H \times W}$. Specifically, the dynamic filtering layer in the DFN \cite{jia2016dynamic} has two types of instances: \textit{dynamic convolutional layer} (with $n_{F}=1$) and \textit{dynamic local filtering layer} (with $n_{F} = H \cdot W$). The latter is adopted in this paper because it guarantees content-adaptive filtering via applying a specific local filter $\mathcal {F}_\theta ^{(u, v)}$ to the neighborhood centered around each pixel coordinate $(u, v)$ of the input $\mathbf{x}_B$:
%$\mathcal {F}_\theta ^{(i,j)}$ to the neighbor centered around $\mathbf{x}_B(i, j)$ for each position $(i, j)$ of the input $\mathbf{x}_B$:

\begin{equation}\label{eq:dfn}
G(u, v) = \mathcal {F}_\theta ^{(u, v)} (\mathbf{x}_B(u, v))
\end{equation}

%\begin{equation}\label{eq:dfn}
%G(i, j) = \mathcal {F}_\theta ^{(i, j)} (\mathbf{x}_B(i, j))
%\end{equation}

\noindent Therefore, the operations in Eq. \ref{eq:dfn} are not only input content specific but also spatial position specific.

%%%%%%%%%%%%%%%%%%%%%%%%%%%%%%%%%%%%%%%%%%
%%%%%%%%%%%%%%%%%%  PAC  %%%%%%%%%%%%%%%%%
%%%%%%%%%%%%%%%%%%%%%%%%%%%%%%%%%%%%%%%%%%
\subsubsection{3.2.3 Pixel Adaptive Convolutional (PAC) Module}\label{subsec:pac}
\begin{figure*}
\begin{center}
\includegraphics[width=0.8\linewidth]{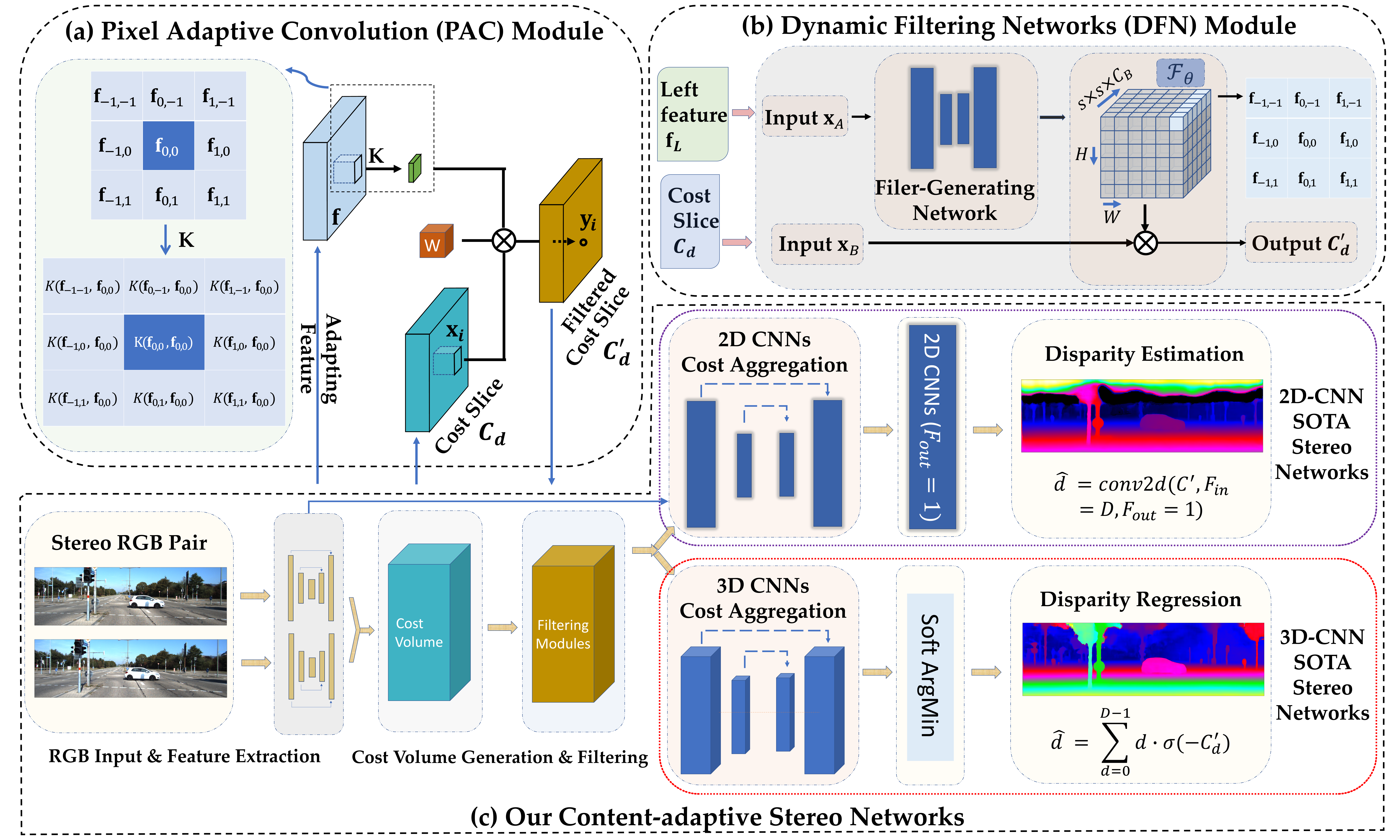}
\end{center}
\vspace{-12pt}
\caption{ (a) %Integrating the PAC module. 
PAC filters the cost volume slice $\mathcal{C}_d$ by multiplying it with a standard spatially invariant convolutional filter $\mathbf{W}$ and a spatially varying filter $K$ that depends on input pixel features $\mathbf{f}$. The output is the filtered slice $\mathcal{C}^{\prime}_d$. 
(b) %Integrating the DFN Module. 
DFN filters $\mathcal {F}_\theta$ are dynamically generated by the \textit{filter-generating network} conditioned on left pixel features $\mathbf{f}_L$. $\mathcal{C}_d$ is filtered by $\mathcal {F}_\theta$ to output $\mathcal{C}^{\prime}_d$. 
(c) Overview of network architecture: a stereo pair is fed into a weight-sharing Siamese sub-network for feature learning; the extracted features of each view are correlated or concatenated to generate the cost volume $\mathcal{C}$; content-adaptive filtering modules %(including SABF, DFN, PAC and SGA) 
are applied to $\mathcal{C}$ and output the filtered volume $\mathcal{C}^{\prime}_d$, followed by state-of-the-art encoder-decoder cost aggregation and disparity regression, as shown in two branches for 2D and 3D convolutional architectures.
}
\label{fig:our-net}
\vspace{-8pt}
\end{figure*}

Pixel adaptive convolution (PAC) proposed by Su \etal \cite{su2019pixel} is a new content-adaptive convolution, which can alleviate the drawback of standard convolution that ignores local image content, while retaining its favorable spatial sharing property compared with existing content-adaptive filters, \eg the DFN \cite{jia2016dynamic}. As illustrated in Fig. \ref{fig:our-net}(a), PAC modifies a conventional spatially invariant convolution filter $\mathbf{W}$ at each position by multiplying it with a position-specific filter $K^{pac}$, the \textit{adapting kernel}. $K^{pac}$ has a pre-defined form, \eg Gaussian: $e^{-\frac{1}{2} || \mathbf{f}_i  - \mathbf{f}_j ||^2}$, where $\mathbf{f}_i$ and $\mathbf{f}_j$ are the \textit{adapting features}, corresponding to pixels $\mathbf{p}_i$ and $\mathbf{p}_j$. The adapting features $\mathbf{f}$ can be either hand-crafted (\eg position and color features) % $\mathbf{f}=(x,y,r,g,b)$) 
or deep features. %Due to the stereo matching scenario in our paper,  
We use deep features extracted from the left image as the adapting features $\mathbf{f}$. 

Given an input $\mathbf{x}_i \in \mathbb{R}^{C_x}$ at pixel $\mathbf{p}_i$, the output $\mathbf{y}_i \in \mathbb{R}^{C_y}$ filtered by PAC is defined as 

\begin{equation}\label{eq:pac}
\mathbf{y}_i = \sum_{j \in \Omega(i)} \, K^{pac}{(\mathbf{f}_i,\mathbf{f}_j)} \,  \mathbf{W}{[\mathbf{p}_i  - \mathbf{p}_j]} \, \mathbf{x}_i  \, + \mathbf{b}
\end{equation}

\noindent where $C_x$ and $C_y$ denote the feature dimension of $\mathbf{x}_i$ and $\mathbf{y}_i$, respectively. $\Omega(\cdot)$ defines an $s \times s$ convolution window. $\mathbf{W} \in \mathbb{R}^{C_y \times C_x \times s \times s}$ and $\mathbf{b} \in \mathbb{R}^{C_y}$ indicate the convolution filter weights and biases. We adopt the notation $[\mathbf{p}_i - \mathbf{p}_j]$ to spatially index filter weights $\mathbf{W}$ with 2D spatial offsets. In our approach $\mathbf{x}_i$ and $\mathbf{y}_i$ are from  the raw and filtered cost volume slices $\mathcal{C}_d$ and $\mathcal{C}^{\prime}_d$, respectively.

%%%%%%%%%%%%%%%%%%%%%%%%%%%%%%%%%%%%%%%%%%
%%%%%%%%%%%%%%%%%%  SGA  %%%%%%%%%%%%%%%%%
%%%%%%%%%%%%%%%%%%%%%%%%%%%%%%%%%%%%%%%%%%
\begin{figure}[t]
\begin{center}
\includegraphics[width=.9\linewidth]{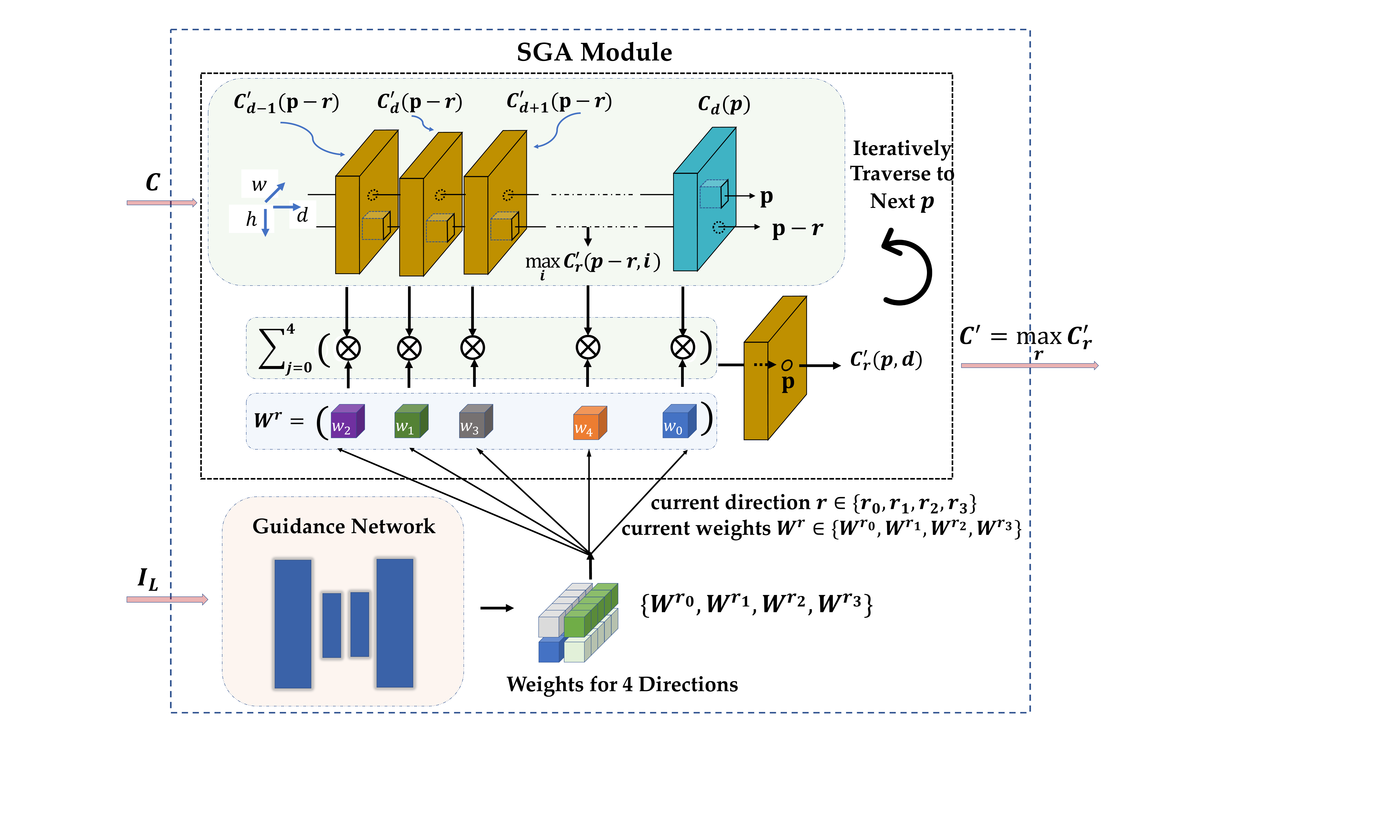}
\end{center}
\vspace{-12pt}
\caption{Integrating the SGA module. The bottom branch shows the guidance network which generates aggregation weights in four directions. %, given the reference image $\mathbf{I}_L$. 
The top branch illustrates SGA which iteratively aggregates the cost volume $\mathcal{C}$ via traversing from pixel $\mathbf{p}-\mathbf{r}$ to $\mathbf{p}$ in a direction $\mathbf{r}$, over the entire image and each disparity $d$. The maximum response among the four directions is selected as the output $\mathcal{C}^{\prime}$.
}
\vspace{-8pt}
\label{fig:sga}
\end{figure}

\subsubsection{3.2.4 Semi-Global Aggregation Module} \label{subsec:sga}
In contrast to the above filters that leverage local context, Semi-Global Aggregation (SGA)  \cite{zhang2019ga} is able to iteratively aggregate the cost volume considering both pixelwise costs and pairwise smoothness constraints in four directions. The constraints are originally defined and approximately solved by Semi-Global Matching (SGM) \cite{hirschmuller08} as an energy function of the disparity map. SGA learns to aggregate the cost volume $\mathcal{C}$ to approximately minimize the energy, and also support backpropagation for end-to-end stereo matching as shown in Fig. \ref{fig:sga}. The aggregated cost volume $\mathcal{C}^{\prime}$ is recursively defined as:

\vspace{-12pt}
\begin{equation}\label{eq:sga}
\begin{split}
\mathcal{C}^{\prime}_{\mathbf{r}}(\mathbf{p}, d)  &= \text{sum } \begin{cases}
 \mathbf{w}_0(\mathbf{p},\mathbf{r}) \cdot \mathcal{C}(\mathbf{p}, d) \\
 \mathbf{w}_1(\mathbf{p},\mathbf{r}) \cdot \mathcal{C}^{\prime}_{\mathbf{r}}(\mathbf{p} - \mathbf{r}, d) \\
 \mathbf{w}_2(\mathbf{p},\mathbf{r}) \cdot \mathcal{C}^{\prime}_{\mathbf{r}}(\mathbf{p} - \mathbf{r}, d-1) \\
 \mathbf{w}_3(\mathbf{p},\mathbf{r}) \cdot \mathcal{C}^{\prime}_{\mathbf{r}}(\mathbf{p} - \mathbf{r}, d+1) \\
 \mathbf{w}_4(\mathbf{p},\mathbf{r}) \cdot \underset{i}{\mathrm{max}} \,\, \mathcal{C}^{\prime}_{\mathbf{r}}(\mathbf{p} - \mathbf{r}, i) \\
 \end{cases} 
 %\\ &\text{ s.t.} \quad \sum_{j=0}^4  \mathbf{w}_j(\mathbf{p},\mathbf{r}) = 1
\end{split}
\end{equation}

%\noindent where, $ \mathbf{r}$ is a unit direction vector in which the cost $\mathcal{C}^{\prime}_{\mathbf{r}}(\mathbf{p}, d)$ of a location $\mathbf{p}$ at disparity $d \in [0, D)$ aggregate along a path. The weights $\mathbf{w}_j$ are achieved and normalized (s.t. $\sum_{j=0}^4  \mathbf{w}_j(\mathbf{p},\mathbf{r}) = 1$) by a guidance sub-network (see Fig.) to avoid very large accumulated values when traversing along the path. The final aggregated cost $\mathcal{C}^{\prime}_{\mathbf{r}}(\mathbf{p}, d)$ is obtained by picking the maximum among four directions of left, right, up and down, i.e., $\mathbf {r} \in \{ (0,1)^\intercal, (0,−1)^\intercal, (1,0)^\intercal, (−1,0)^\intercal \}$, as defined:

\noindent where $ \mathbf{r}$ is a unit direction vector along which the cost $\mathcal{C}^{\prime}_{\mathbf{r}}(\mathbf{p}, d)$ of pixel $\mathbf{p}$ at disparity %$d \in [0, D)$ 
$d$ is aggregated. % along a path. PM: already said direction vector
The weights $\mathbf{w}_j$ are achieved and normalized (s.t. $\sum_{j=0}^4  \mathbf{w}_j(\mathbf{p},\mathbf{r})=1$) by a guidance sub-network to avoid very large accumulated values when traversing along the path. The final aggregated cost $\mathcal{C}^{\prime}_{\mathbf{r}}(\mathbf{p}, d)$ is obtained by picking the maximum among four directions, namely, left, right, up and down, i.e., $\mathbf{r} \in \{ (-1,0), (1,0), (0,1), (0,-1) \}$, defined as:

\begin{equation} \label{eq:sga-final}
\mathcal{C}^{\prime}(\mathbf{p}, d) = \underset{\mathbf{r}}{\mathrm{max}} \,\, \mathcal{C}^{\prime}_{\mathbf{r}}(\mathbf{p}, d)
\end{equation}

%\noindent The backpropagation of $\mathbf{w}_j$ and $\mathcal{C}(\mathbf{p}, d)$ in the SGA layer can be done inversely as Eq. \ref{eq:sga}. 

\subsection{Network Architecture}\label{sec:sub-network}

As illustrated in Fig. \ref{fig:our-net}, our architecture takes as backbones four state-of-the-art 2D and 3D CNNs for stereo matching, i.e., DispNetC \cite{mayer2016large}, GCNet \cite{kendall2017-gcnet}, PSMNet \cite{chang2018psmnet} and GANet \cite{zhang2019ga}, and adapts %four effective content-adaptive filtering techniques as neural network layers, including 
SABF  (Fig. \ref{fig:sabf}), PAC (Fig. \ref{fig:our-net}(a)), DFN (Fig. \ref{fig:our-net}(b)) and SGA  (Fig. \ref{fig:sga}), to aggregate the cost volume. % for end-to-end stereo matching. 

%\subsubsection{3.3.1 Network Backbones and Capacities} \label{sec:backbones}
\subsubsection{3.3.1 Network Backbones} \label{sec:backbones}

Given a rectified stereo pair, % with dimension $H \times W$ and the maximum disparity $D$, 
the backbone architectures described below include unary feature extraction from the images by a weight-sharing Siamese structure, cost volume computation and regularization, and disparity regression. Unary features are denoted by $\mathbf{f}_u$. 2D or 3D convolutional and transpose convolutional layers have $3 \times 3$ or $3 \times 3 \times 3$ kernels, unless otherwise specified. %For convenience unary features are denoted by $\mathbf{f}_u$. %Unless otherwise specified, each convolutional layer is followed by a batch normalization layer and a ReLU layer.

\vspace{3pt} \noindent \textbf{DispNetC Backbone.} \, DispNetC \cite{mayer2016large} is an encoder-decoder architecture with an explicit 1D correlation layer. The encoder downsamples the input images via convolution by up to $1/64$, while the decoder gradually upsamples the feature maps via transposed convolution at 6 different scales ranging from $1/64$ to $1/2$. 
%%%%%%%%%%%%%%%%%%%%%%%%%%%%%%%%%%%%%%%
%% update for camera-ready version
%The input images are processed by a Siamese network (layers of {\myqcrfont conv1} and {\myqcrfont conv2}) to generate $\mathbf{f}_u$ ($H/4 \times W/4 \times F$, $F=128$). Features $\mathbf{f}_u$ are correlated by a 1D correlation layer \cite{dosovitskiy2015flownet} to form a 3D cost volume $\mathcal{C}$ ($D/4 \times H/4 \times W/4$) with a maximum disparity of $D/4$. 
%The volume is further contracted to $1/64$ (layer {\myqcrfont conv6} and {\myqcrfont loss6}) resolution and expanded by alternate transpose convolutions ({\myqcrfont upconv}\textit{N}), convolutions ({\myqcrfont iconv}\textit{N}, {pr}\textit{N}) and loss layers ({\myqcrfont loss}\textit{N}), where $N=5, 4, \dots, 1$. There are six intermediate disparity maps which are all interpolated to the input resolution. The loss during training is computed on all six disparity maps, but only the last one is used as output.
Features $\mathbf{f}_u$ ($H/4 \times W/4 \times F$, $F=128$) are correlated by a 1D correlation layer \cite{dosovitskiy2015flownet} to form a 3D cost volume $\mathcal{C}$ ($D/4 \times H/4 \times W/4$) with a maximum disparity of $D/4$.
$\mathcal{C}$ is further contracted to $1/64$ resolution and expanded by alternate transpose convolutions and loss layers. There are six intermediate disparity maps which are all interpolated to the input resolution. The loss is computed on all six disparity maps, but only the last one is used as output.

\vspace{3pt} \noindent \textbf{GCNet Backbone.} \, GCNet \cite{kendall2017-gcnet} is a 3D CNN architecture that models geometry in stereo matching. 
We add an initial bilinear interpolation layer
%, which has no parameters and helps reduce the computational burden, 
to downsample the input stereo pair to half resolution.
GCNet extracts $\mathbf{f}_u$ ($F \times H/4 \times W/4$, $F=32$) via a $5 \times 5$ convolution layer and eight residual blocks \cite{he2016_resnet} and then builds a 4D cost volume $\mathcal{C}$ ($D/4 \times 2F \times H/4 \times W/4$) by concatenating $\mathbf{f}_u$ with its counterpart from the other view across each disparity level. $\mathcal{C}$ is downsampled by up to $1/16$ via four encoding blocks (each with three convolutions with strides equal to $2$, $1$, and $1$) and upsampled by $32$ via five decoding blocks (each has a skip-connection from the early layer and one transposed convolution with stride $2$), resulting in an aggregated ($D/2 \times H/2 \times W/2$) volume. It is interpolated to input resolution and regressed to predict the disparity map via the differentiable \textit{soft argmin}: % (Eq. \ref{eq:softargmin}):

\begin{equation}\label{eq:softargmin}
%\hat{d} = \sum_{d=0}^{D-1} d \times \sigma(- \mathcal{C}_d)
\hat{\mathcal{D}} = \sum_{d=0}^{D-1} d \cdot \sigma(- \mathcal{C}_d)
\end{equation} 

\noindent where cost $\mathcal{C}_d$
%for a given pixel $\mathbf{p}$ 
is first converted to a probability of each disparity value $d$ % \in [0, D)$ 
via the \textit{softmax} operation $\mathbf{\sigma} (\cdot)$.The disparity map $\hat{\mathcal{D}}$ comprises the \textit{expected} values of $d$ for each pixel.

\vspace{3pt} \noindent \textbf{PSMNet Backbone.} \, PSMNet \cite{chang2018psmnet} learns $\mathbf{f}_u$ ($F \times H/4 \times W/4$, $F=128$) via three convolution layers followed by four basic residual blocks \cite{he2016_resnet}. 
%%%%%%%%%%%%%%%%%%%%%%%%%%%%%%%%%%%%%%%
%% update for camera-ready version
%To incorporate hierarchical global context, 
$\mathbf{f}_u$ is further processed separately by spatial pyramid pooling (SPP), a $1 \times 1$ convolution, bilinear interpolation and feature concatenation. This generates the SPP features ($F \times H/4 \times W/4$, $F=32$) used to construct a 4D  ($D/4 \times 2F \times H/4 \times W/4$) cost volume $\mathcal{C}$. $\mathcal{C}$ is regularized by a stacked hourglass %architecture 
with three encoding-decoding blocks. %each with four  convolution layers (with strides $2$, $1$, $2$ and $1$), and two transposed convolution layers to restore the input resolution. 
Each hourglass generates a regularized volume, from which an intermediate disparity is obtained via the \textit{soft argmin} (Eq. \ref{eq:softargmin}). The training loss is computed on all three intermediate disparities.
%but only the last one is output at test time.

\vspace{3pt} \noindent \textbf{GANet Backbone.} \, GANet \cite{zhang2019ga} introduces local guided aggregation (LGA) and semi-global aggregation (SGA) layers  
which are complementary to 3D convolutional layers. The unary features $\mathbf{f}_u$ ($F \times H/3 \times W/3$, $F=32$) are extracted through a stacked hourglass CNN 
with skip connections. They are then used to construct a 4D cost volume ($D/3 \times 2F \times H/3 \times W/3$), which is fed into a cost aggregation block (built of alternate 3D convolution and transpose convolution layers, SGA and LGA) for regularization, refinement and disparity regression via the \textit{soft argmin} (Eq. \ref{eq:softargmin}). The guidance sub-network generates the weight matrices for SGA (Sec. \ref{subsec:sga}) and LGA. The LGA layer is used before disparity regression and locally refines the 4D cost volume for several times. We adopt the GANet-deep version which achieves the best accuracy among all variants. %(\url{https://github.com/feihuzhang/GANet})

\vspace{3pt} \noindent \textbf{Network Capacity.} \,Network capacities are listed in Table \ref{tab:net-params}. The parameters of four backbone networks (i.e, row W/O in gray, meaning no filters applied) increase in the order of GCNet $<$ PSMNet $<$ GANet $<$ DispNetC. The order of the filtering modules (rows in skyblue) %have different complexities in terms of parameter increase percentage  when incorporated into those backbones, shown from smallest to largest as 
according to increasing number of parameters is PAC $<$ DFN $<$ SABF $<$ SGA. 
%Their performance are carefully investigated in Sec. \ref{sec:experiments}. 

\subsubsection{3.3.2 Loss Function}
%and Network Capacity}
\label{sec:sub-loss-netcapacity}
%\subsubsection{3.3.2 Loss Function and Weights}\label{sec:sub-loss}

%The loss is evaluated over the valid pixels which have ground truth disparity. 

%\begin{equation}\label{eq:l1_loss_smooth}
%\begin{aligned}[b]
%L(d, \hat{d}) & = \frac{1}{N} \sum_{i=1}^N l_s(|| d_i - \hat{d_i} ||_1) \\
%\text{and, } l_s(x) &=  \begin{cases} 0.5x^2 & \text{if } |x| < 1 \\ |x| - 0.5 & \text{otherwise}
%\end{cases}
%\end{aligned}
%\end{equation}

%\noindent where $N$ counts the valid pixels, $|| d_i - \hat{d_i} ||_1$ measures the absolute error of disparity prediction $\hat{d_i}$ and ground truth $d_i$. 

%\noindent \textbf{Loss Function and Weights.} \, 
The smooth $L_1$ loss function (see the supplement) is used for end-to-end training. The GCNet backbone has one disparity output, while the other backbones produce multiple intermediate disparity maps. The network loss is their weighted average.
%%Their weights for each \PMcomments{scale???} (\ccjscomments{for each disparity output, since they have been interpolated to the image resolution}) are defined as $0.2$, $0.6$ and $1.0$ for GANet and $0.5$, $0.7$ and $1.0$ for PSMNet. DispNetC training starts on Scene Flow with weights of $0.05, 0.10, 0.14, 0.19, 0.24$ and $0.29$ for the six disparity estimates. In finetuning, only the final disparity is activated. When integrating the SABF filter to the backbones, the embedding loss in Eq. \ref{eq:embeddingLoss} is added with a weight of $0.06$. 
The corresponding weights are defined as 1) GANet: $0.2$, $0.6$, and $1.0$; 2) PSMNet: $0.5$, $0.7$, and $1.0$; 3) DispNetC: $0.05$, $0.10$, $0.14$, $0.19$, $0.24$, and $0.29$ as training starts on Scene Flow, while in finetuning, only the final disparity is activated. When integrating the SABF filter to the backbones, the embedding loss in Eq. \ref{eq:embeddingLoss} is added with a weight of $0.06$.

%\subsubsection{3.3.3 Network Capacity}\label{sec:sub-netcapacity}
% \vspace{1pt} \noindent \textbf{Network Capacity.} \,Network capacities are listed in Table \ref{tab:net-params}. The parameters of four backbone networks (i.e, row W/O in gray, meaning no filters applied) increase in the order of GCNet $<$ PSMNet $<$ GANet $<$ DispNetC. The order of the filtering modules (rows in skyblue) %have different complexities in terms of parameter increase percentage  when incorporated into those backbones, shown from smallest to largest as 
% according to increasing number of parameters is PAC $<$ DFN $<$ SABF $<$ SGA. 
% %Their performance are carefully investigated in Sec. \ref{sec:experiments}. 

\begin{table}[t]
\small
\centering
\scalebox{0.9}{
    %\begin{tabularx}{\linewidth}{p{0.75cm} | XX XX XX XX}
    \begin{tabularx}{\linewidth}{ @{} ?{1pt} p{0.75cm} ?{1pt} YY ?{1pt} YY ?{1pt} YY ?{1pt} YY ?{1pt} @{} }
    %\hline
    \Xhline{2\arrayrulewidth}
	\multirow{3}{0.8pt}{{\bf Filters}} & \multicolumn{8}{c ?{1pt} }{\bf Network Backbones}\\
	\cline{2-9} % draw a horizontal line spanning only some of the table cells
	& \multicolumn{2}{c ?{1pt} }{\bf DispNetC} & \multicolumn{2}{c ?{1pt} }{ \bf PSMNet}  & \multicolumn{2}{c ?{1pt}}{ \bf GANet} & \multicolumn{2}{c ?{1pt}}{ \bf GCNet} \\
	\cline{2-9} % draw a horizontal line spanning only some of the table cells
	 &{No.} & {$\uparrow$(\%)} &{No.} & {$\uparrow$(\%)} &{No.} & {$\uparrow$(\%)} &{No.} & {$\uparrow$(\%)}  \\
	%\hline
	\Xhline{2\arrayrulewidth}
	\cellcolor{mygray}W/O & 42.2 & - &  5.2  & - & 6.6 & -  & 2.8 & - \\
	\Xhline{2\arrayrulewidth}
	\cellcolor{myskyblue}SABF & 44.0 & 4.2 & 7.0 & 34 & \textbf{8.4} & \textbf{27}  & 4.6 & 62.4 \\
	\cellcolor{myskyblue}DFN & 42.6 & 0.8 & 5.6 & 6.4 & 6.9 & 5.1  & 3.2 & 11.8 \\
	\cellcolor{myskyblue}PAC & 42.3 & 0.1 & 5.3 & 2.0 & 6.7 & 1.6  & 2.9 & 3.6 \\
	\cellcolor{myskyblue}SGA & \textbf{45.2} & \textbf{7.0} & \textbf{8.3} & \textbf{58.8} & - & - & \textbf{5.9} & \textbf{108} \\
	%\hline
	\Xhline{2\arrayrulewidth}
	\end{tabularx}}
\vspace{-4pt}
\caption{Network capacity. For each combination of the filtering techniques and network backbones, columns \textit{No.} show the number of parameters in millions, and columns $\uparrow$(\%) are the relative increase in the number of parameter w.r.t. the backbone baselines. Largest values are in bold. Inapplicable entries are marked by ``-''.}
\label{tab:net-params}
\end{table}
%\vspace{-12pt}

%\subsubsection{3.3.2 Disparity Regression} \label{sec:dispRegress}
%There are two distinctive paradigms of disparity regression in end-to-end stereo matching: i) applying 2D convolution over the 3D cost volume in 2D convolutional architectures; ii) applying \textit{soft argmin} to the 4D cost volume in 3D convolutional architectures.

%\vspace{3pt} \noindent \textbf{Disparity estimation in 2D CNNs.} \, This category of networks applies a 2D convolution over the 3D cost volume (with dimensions $D\times H \times W$) by setting  {\myqcrfont in\_channels=D} and {\myqcrfont out\_channels=1}, resulting in the disparity map $\mathcal{D}$ with dimensions $H \times W$ as output.

%\begin{equation}\label{eq:disp_2DCNNs}
%\mathcal{D}(u,v) = \textit{Conv2d}(\mathcal{C}(u,v), C_{in}=D, C_{out}=1))
%\end{equation}

%\vspace{3pt} \noindent \textbf{Disparity regression in 3D CNNs.} \, State-of-the-art 3D-CNN stereo networks employ the differentiable \textit{soft argmin} \cite{kendall2017-gcnet} on the 4D cost volume for disparity regression. Specifically, the cost curve $\mathcal{C}^{\prime}_d$ for a given pixel $\mathbf{p}$ is first converted to a probability of each disparity $d \in [0, D)$ using the \textit{softmax} operation $\mathbf{\sigma} (\cdot)$. Then the \textit{expected} value of $d$ is calculated as the regressed disparity $\hat{d}$:
%\begin{equation}\label{eq:softargmin} \hat{d} = \sum_{d=0}^{D-1} d \times \sigma(- \mathcal{C}^{\prime}_d) \end{equation} 

\section{Experiments}\label{sec:experiments}

% -----------------------------
% --- Paper in Review Version
% -----------------------------

%\input{experiments}

% -----------------------------
% --- Paper in Camera-ready Version
% -----------------------------
%\vspace{4pt} \noindent  \textbf{Datasets} \,\, 
\subsection{Datasets}

Our networks are trained from scratch on Scene Flow \cite{mayer2016large}, then finetuned on Virtual KITTI 2 (VKT2) \cite{cabon2020virtual,gaidon2016virtual} or KITTI 2015 (KT15) \cite{menze2015object}. \textbf{Scene Flow} is a large scale synthetic dataset containing $35,454$ training and $4,370$ testing images with dense ground truth disparity maps.
%with dimensions $540 \times 960$. 
We exclude the pixels with disparities $d>192$ in training.
\textbf{Virtual KITTI 2 (VKT2)} is a synthetic clone of the real KITTI dataset. It contains 5 sequence \textit{clones} of {\myqcrfont Scene 01, 02, 06, 18} and {\myqcrfont 20}, and nine variants with diverse weather conditions (e.g. {\myqcrfont fog, rain}) or modified camera configurations (e.g. rotated by $15 \degree$, $30 \degree$). Since there is no designated validation set, we select (i) {\myqcrfont Scene06} (i.e., \textit{VKT2-Val-S6} with $2,700$ frames) and (ii) multiple blocks of consecutive frames from {\myqcrfont Scene 01, 02, 18, 20} (i.e., \textit{VKT2-Val-WoS6} with $2,620$ frames), and use the remaining $15,940$ images for training. The former scene remains unseen during training, while networks observe similar frames to the latter. We evaluate these settings separately.
\textbf{KITTI 2015 (KT15):} is a real dataset of street views.
% from a driving car. 
It contains $200$ training stereo image pairs with sparsely labeled disparity from LiDAR data. We divided the training data into a training set ($170$ images) and a validation set ($30$ images) for our experiments.

\noindent \textbf{Metrics.} \, For KT15, we adopt the \textit{bad-3} error
(i.e., percentage of pixels with disparity error $>3$px or $\geq 5\%$ of the true disparity) counted over non-occluded (noc) or all pixels with ground truth. For VKT2, in addition to \textit{bad-3}, we also use the endpoint error (EPE) and \textit{bad-1} error ($\geq 1$px or $\geq 5\%$) over all pixels.

\begin{table*}[t]
\small
\centering
\scalebox{0.9}{
    %\begin{tabularx}{1.0\linewidth}{p{0.75cm} | XX XX XX XX}
    %\begin{tabularx}{1.0\linewidth}{@{} p{0.75cm} | YYY| YYY |YYY | YYY @{}}
    \begin{tabularx}{1.0\linewidth}{@{} ?{1pt} p{0.75cm} ?{1pt} YYY ?{1pt} YYY  ?{1pt}YYY ?{1pt} YYY ?{1pt} @{}}
    %\hline
    \Xhline{2\arrayrulewidth}
	\multirow{3}{0.8pt}{{\bf Filters}} & \multicolumn{3}{c ?{1pt} }{ \bf DispNetC} & \multicolumn{3}{c ?{1pt}  }{ \bf PSMNet}  & \multicolumn{3}{c ?{1pt}  }{ \bf GANet} & \multicolumn{3}{c ?{1pt}}{ \bf GCNet} \\
	\cline{2-13} % draw a horizontal line spanning only some of the table cells
	%&\multicolumn{8}{c}{Bad-3.0 Error Rates (\%)}  \\
	 &{EPE(px)}  &{$\geq 1$ px} &{$\geq 3$ px} &{EPE(px)}  &{$\geq 1$ px} &{$\geq 3$ px} &{EPE(px)}  &{$\geq 1$ px} &{$\geq 3$ px} &{EPE(px)}  &{$\geq 1$ px} &{$\geq 3$ px} \\
	%\hline \hline
	\Xhline{2\arrayrulewidth}
	\multicolumn{13}{ ?{1pt} c ?{1pt}  }{ \textit{Seen locations (i.e., Scene01, 02, 18 and 20) from VKT2 validation set \textbf{VKT2-Val-WoS6}}} \\
	\hline
	W/O & 0.68 & 11.54 & 3.72  &0.45   & 7.08  & 2.21  & \textbf{0.33}  & \textbf{5.34} & \textbf{1.70}  & 0.62  & 9.71 & 3.18 \\
	\hline
	%\Xhline{2\arrayrulewidth}
	\cellcolor{myskyblue}SABF & \cellcolor{mygray}0.65 & \cellcolor{mygray} 10.86 & \cellcolor{mygray} 3.48  & \cellcolor{mygray}\textbf{0.36} & \cellcolor{mygray}\textbf{5.92} & \cellcolor{mygray}\textbf{1.98}  & 0.34 & 5.50 & 1.76  & \cellcolor{mygray}0.60 & 9.89  & 3.19  \\
	\cellcolor{myskyblue} DFN & \cellcolor{mygray} \textbf{0.57} & \cellcolor{mygray}9.85 & \cellcolor{mygray}3.26  & \cellcolor{mygray}0.42  & \cellcolor{mygray}6.45 & \cellcolor{mygray}2.17  & 0.37  & 6.23 & 1.99  & \cellcolor{mygray}0.60 & \cellcolor{mygray}9.20 & \cellcolor{mygray}3.11  \\
	\cellcolor{myskyblue}PAC & \cellcolor{mygray}0.58 & \cellcolor{mygray}9.92 & \cellcolor{mygray}3.39  & 0.52  & 7.81  & 2.61   & 0.40  & 7.01 & 2.20  & 0.75 & 12.98 & 4.02  \\
	\cellcolor{myskyblue}SGA & \cellcolor{mygray}\textbf{0.57} & \cellcolor{mygray}\textbf{9.37} & \cellcolor{mygray}\textbf{3.21} & \cellcolor{mygray}0.40 & \cellcolor{mygray}6.08 & \cellcolor{mygray}2.14  &- &- &-  & \cellcolor{mygray}\textbf{0.55} & \cellcolor{mygray}\textbf{9.24} & \cellcolor{mygray}\textbf{2.98}  \\
	%\hline
	\Xhline{2\arrayrulewidth}
	
    \multicolumn{13}{ ?{1pt} c ?{1pt}  }{ \textit{Totally unseen location (i.e., Scene06) from VKT2 validation set \textbf{VKT2-Val-S6}}} \\
	\hline
	%\Xhline{2\arrayrulewidth}
	W/O & 0.70 & 10.28 & 3.12  & 0.48   & 5.16  & 1.96  & 0.30  & \textbf{3.09} & 1.0563  & 0.59  & 7.48 & 2.25 \\
	\hline
	%\Xhline{2\arrayrulewidth}
	\cellcolor{myskyblue}SABF & \cellcolor{mygray}0.69 & \cellcolor{mygray} 9.75 & \cellcolor{mygray} 3.00  & \cellcolor{mygray}0.44 & \cellcolor{mygray}4.47 & \cellcolor{mygray}1.73  & \cellcolor{mygray}\textbf{0.28} & 3.16 & \cellcolor{mygray}\textbf{0.97}  & \cellcolor{mygray}0.56 & 7.51  & \cellcolor{mygray}2.23  \\
	\cellcolor{myskyblue} DFN & \cellcolor{mygray} \textbf{0.599} & \cellcolor{mygray}8.54 & \cellcolor{mygray}\textbf{2.791}  & \cellcolor{mygray}\textbf{0.39}  & \cellcolor{mygray}4.83 & \cellcolor{mygray}\textbf{1.69}  & \cellcolor{mygray}0.29  & 3.54 & \cellcolor{mygray} 1.0561  & \cellcolor{mygray}0.55 & \cellcolor{mygray}6.81 & \cellcolor{mygray}\textbf{2.14}  \\
	\cellcolor{myskyblue}PAC & \cellcolor{mygray}0.603 & \cellcolor{mygray}8.73 & \cellcolor{mygray}2.96  & 0.52  & 5.78  & 1.98   & 0.35  & 4.36 & 1.47  & 0.73 & 11.87 & 2.99  \\
	\cellcolor{myskyblue}SGA & \cellcolor{mygray}0.607 & \cellcolor{mygray}\textbf{8.02} & \cellcolor{mygray}2.794 & \cellcolor{mygray}0.42 & \cellcolor{mygray}\textbf{4.34} & \cellcolor{mygray}1.71  &- &- &-  & \cellcolor{mygray}\textbf{0.53} & \cellcolor{mygray}\textbf{7.45} & 2.29  \\
	%\hline
	\Xhline{2\arrayrulewidth}
	\end{tabularx}}
\vspace{-4pt}
\caption{Evaluation on the validation sets of VKT2-Val-WoS6 and VKT2-Val-S6. 
%The error metric $\geq$ 1 (or $3$) px shows the ratio of bad pixels with estimation error $\geq 1$ (or $3$) pixels or $\geq 5$\% of the true disparity. 
Results are shown in the entries of filters (rows in skyblue) and backbones (columns) w.r.t. the baselines (rows W/O). Our improved results are highlighted in gray, and the best ones are in bold. GANet already contains SGA, resulting in blank entries ``-''.
}
\label{tab:vkt2-val}
\end{table*}

\iftrue
\begin{figure*}[t]
    \centering
    \scalebox{0.95}{
    \renewcommand{\tabcolsep}{0.2pt}
    \scriptsize
    \begin{tabular}{cccc}
        \includegraphics[width=0.24\textwidth]{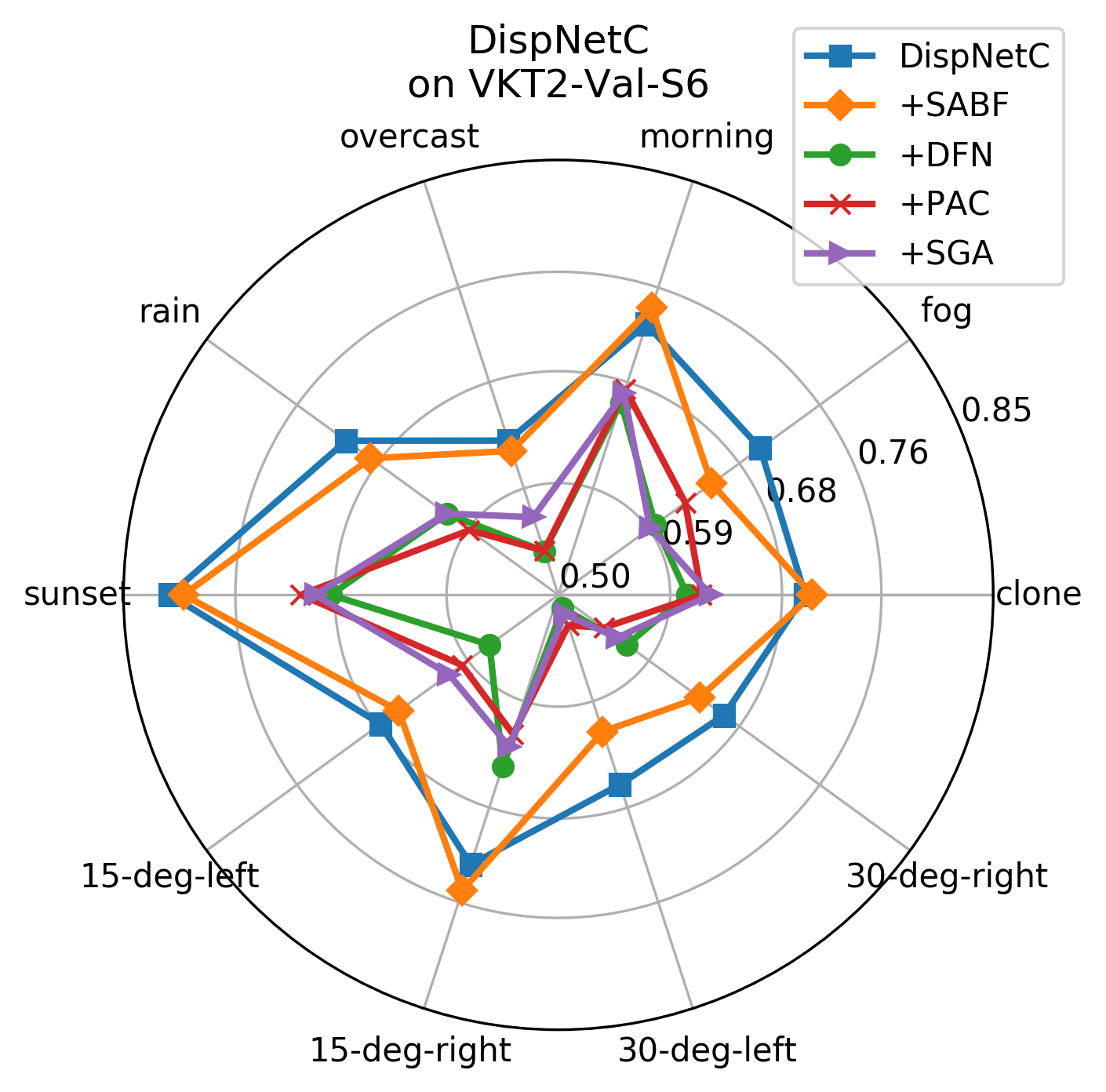} &  \includegraphics[width=0.24\textwidth]{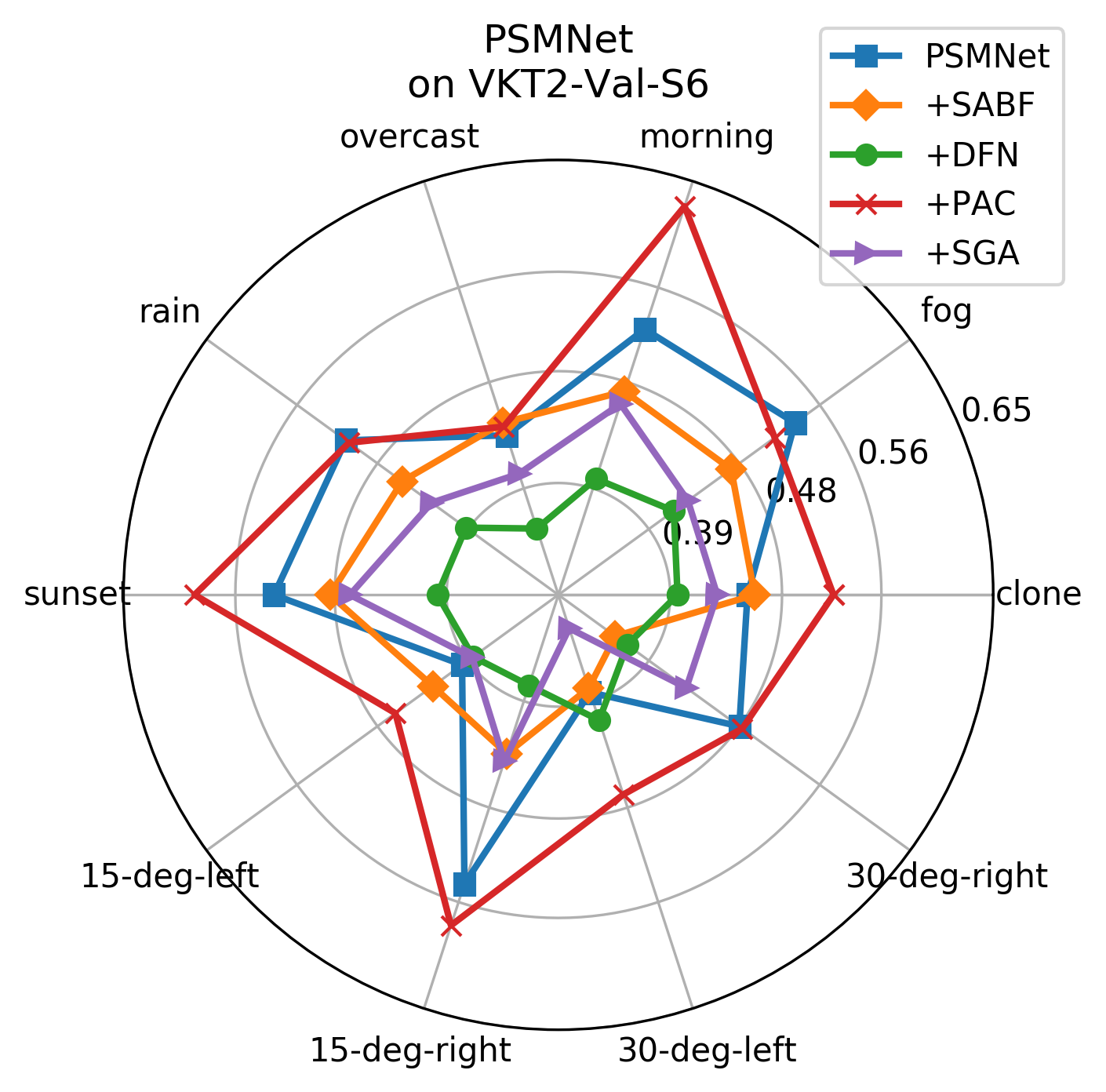} &
        %\normalsize (a) & \normalsize(b) \\ 
        \includegraphics[width=0.24\textwidth]{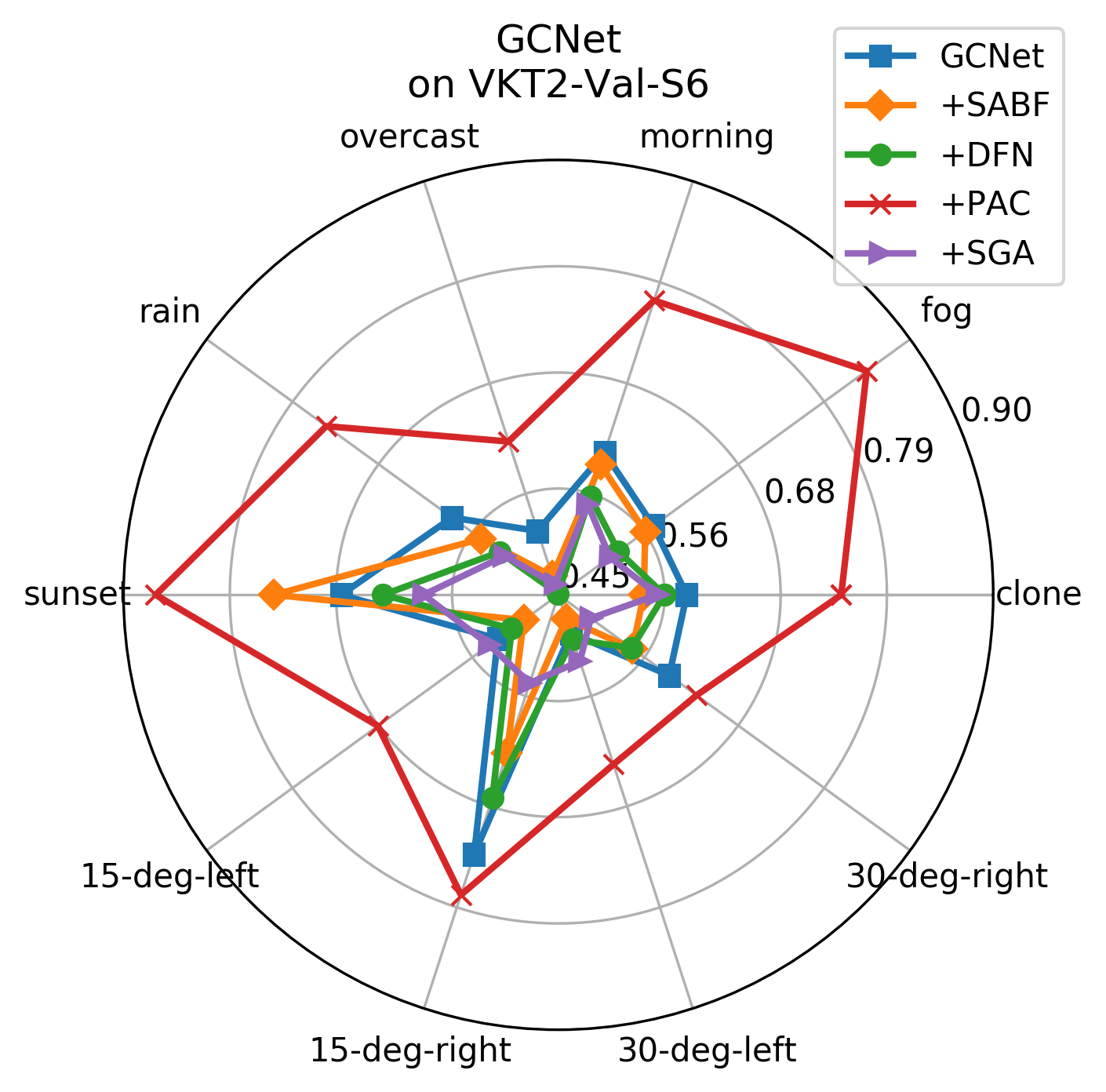} &
        \includegraphics[width=0.24\textwidth]{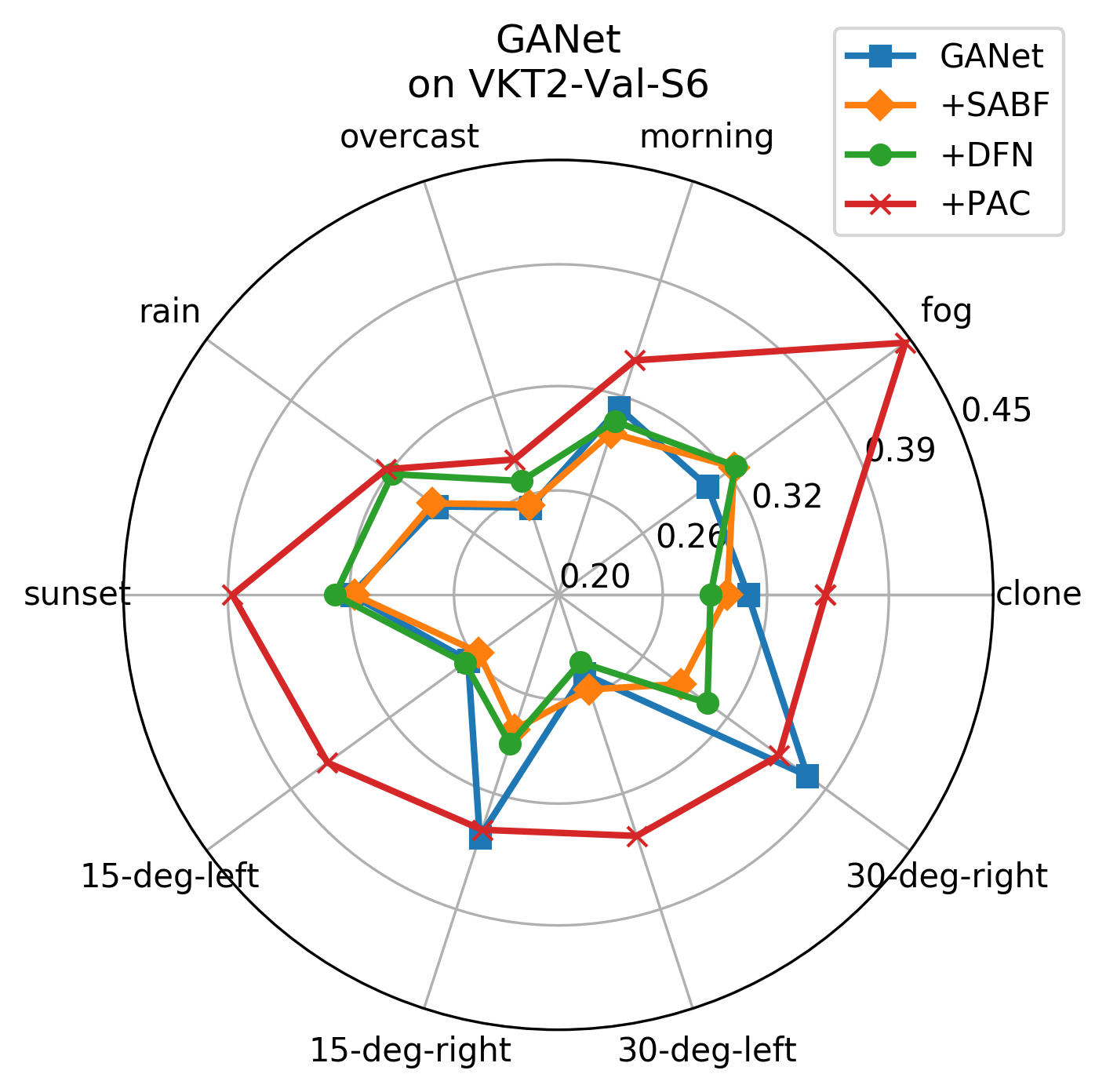} \\
        \normalsize (a) & \normalsize (b) & \normalsize (c)  & \normalsize (d)
    \end{tabular}
    }
    \vspace{-8pt}
    \caption{Results of (a) DispNetC, (b) PSMNet, (c) GCNet and (d) GANet on each category of validation set \textit{VKT2-Val-S6}.}
    \label{fig:vkt2-val-s6}
\end{figure*}
\fi

\iffalse
\begin{figure}[t]
    \centering
    \renewcommand{\tabcolsep}{0.2pt}
    \scriptsize
    \begin{tabular}{cc}
        \includegraphics[width=0.24\textwidth]{img/spider-vkt2-val-s6-dispnetc.png} &  \includegraphics[width=0.24\textwidth]{img/spider-vkt2-val-s6-psmnet.png} \\
        \normalsize (a) & \normalsize(b) \\ 
        \includegraphics[width=0.24\textwidth]{img/spider-vkt2-val-s6-gcnet.png} &
        \includegraphics[width=0.24\textwidth]{img/spider-vkt2-val-s6-ganet.png} \\
        \normalsize (c)  & \normalsize (d)
    \end{tabular}
    \vspace{-8pt}
    \caption{Ours versus (a) DispNetC, (b) PSMNet, (c) GCNet and (d) GANet on each category of the validation set \textbf{VKT2-Val-S6}.}
    \label{fig:vkt2-val-s6}
\end{figure}
\fi

\subsection{Implementation Details}

\vspace{3pt} \noindent \textbf{Architecture Details.} \, Our models are implemented with PyTorch. Each convolution layer is followed by batch normalization (BN) and ReLU unless otherwise specified. We use the official code of PSMNet and GANet, and implement DispNetC (no PyTorch version) and GCNet (no official code). Our implementations achieve similar or better results on the  KT15  benchmark, i.e., bad-3 errors $4.48\%$ (all) and $3.85\%$ (noc) versus the authors' $4.34\%$ (all) and $4.05\%$ (noc) for DispNetC, and $2.38\%$ (all) and $2.07\%$ (noc) versus the authors' $2.87\%$ (all) and $2.61\%$ (noc) for GCNet.
% D1-all error  of  2.76%  (versus  2.87%  as  reported by GCNet).
%There is no BN layer after several convolution layers which output disparity maps in DispNetC and hence in our DispNetC variant. %
We use $\sigma_s=0.7$ and $\sigma_r=0.1$ in Eq. \ref{eq:ebf-filter} for SABF. We investigate how the filter window $s$ and dilation rate $r$ affect the filtering output, and find that $s=5$ with $r=2$ achieve a good balance in accuracy, space requirements and runtime. Please see the supplementary material for ablation studies.

\vspace{3pt} \noindent \textbf{Training Details.} \, The SABF embedding network is pre-trained on Cityscapes \cite{cordts2016cityscapes}. For data augmentation we randomly crop $256 \times 512$ image patches and do channel-wise standardization by subtracting the mean and dividing by the standard deviation. % per channel. 
For fair comparison, we follow the baselines and set the maximum disparity to $D=192$. We use the official pre-trained models on Scene Flow for GANet and PSMNet, and train DispNetC and GCNet on Scene Flow from scratch for $10$ epochs with a learning rate (lr) of $0.001$. Training is optimized with Adam ($\beta_1 = 0.9$, $\beta_2 = 0.999$), except for GCNet which uses RMSprop ($\alpha=0.9$).  For KT15, our models and baselines are finetuned for 600 epochs (with lr of 0.001 for the first $300$ epochs and 0.0001 for the next $300$ epochs). For VKT2, we finetune all algorithms for $20$ epochs, with lr of 0.001 at first, then divided by 10 at epoch 5 and epoch 18.

% solve: Losing the vertical line on my table when using multicolumn 
%Answer: change \multicolumn{11}{l}{...} to \multicolumn{11}{l|}{...} 

% \subsection{Ablation Study}
%  We perform ablation studies to investigate how the filter window $s$ and the dilation rate $r$ can affect the filtering output and disparity estimation. Out of a large number of possible combinations, %we cannot take the ablation study on each of them. 
%  we show \textit{SABF+DispNetC} as a representative. %due to its computational efficiency. 
%  As listed in Table \ref{tab:filter_size_kt15-val30}, we find $s=5$ with $r=2$ achieves a good balance in accuracy, space and runtime. 
%  %The 500-run averaged memory and runtime in GPUs are measured when we test a $384 \times 1280$ stereo pair, and the \textit{bad-3} (noc,all) errors are evaluated on the KITTI 2015 validation set. 
%  Please note that a $5 \times 5$ filter (with dilation rate $2$) covers $9 \times 9$ regions in the cost feature space, which is equivalent to $33 \times 33$ regions in the RGB image space, due to the cost volume being a quarter of the size of the input images. In the following experiments, we keep using $s = 5$ with $r = 2$ for our different architectures.
 
 % Leave a note here: how it is obtained: s = 5, r = 2, (s-1) * r * (img_size / cost_size) + 1 = 33;
 
\subsection{Quantitative Results}
All the results are on the \textit{validation sets} since we could not submit all combinations to the benchmarks. Due to page limits, qualitative results are available in the supplement.

\vspace{3pt} \noindent \textbf{Virtual KITTI 2 Evaluation.} \,We finetune our models (the pre-trained backbones on Scene Flow after integrating the filters) on the VKT2 dataset. (\textit{EPE}, \textit{bad-1} and \textit{bad-3}) on the \textit{VKT2-Val-WoS6} and \textit{VKT2-Val-S6} validation sets are listed in Table \ref{tab:vkt2-val}. In most cases, our models (rows in skyblue) achieve higher accuracy than the baselines (row W/O). The exception is standard GANet which performs well since it always includes SGA and LGA for global and local matching volume aggregation. Fig. \ref{fig:vkt2-val-s6} plots EPE errors on \textit{individual} categories of \textit{VKT2-Val-S6}. %Recall that DispNetC belongs to 2D CNNs and the others belong to 3D CNNs. 
SABF, DFN and SGA  boost 2D and 3D CNNs, while PAC improves 2D CNNs, but not 3D CNNs. When moving from familiar  (\textit{VKT2-Val-WoS6}) to unseen (\textit{VKT2-Val-S6}) validation scenes, DFN achieves better adaptation than SGA, and the SABF and DFN variants outperform standard GANet.

\vspace{3pt} \noindent \textbf{KITTI 2015 Evaluation.} \,The results on KT15 are shown in Table \ref{tab:kt15-val30}. Our networks obtain improved accuracy for all the backbones except for GANet. All the filters boost the backbones due to leveraging image context as guidance, and SGA achieves the highest accuracy among them.  
%\subsection{KITTI 2015 Evaluation}
\begin{table}[b]
\small
\centering
\scalebox{0.8}{
    %\begin{tabularx}{1.0\linewidth}{p{0.75cm} | XX XX XX XX}
    %\begin{tabularx}{1.0\linewidth}{@{} p{0.75cm} | YY| YY| YY| YY @{}}
    %note: ?{1pt} is used to draw a thick vertical line;
    \begin{tabularx}{1.2\linewidth}{@{} ?{1pt} p{0.75cm}?{1pt}  YY ?{1pt}  YY ?{1pt} YY ?{1pt} YY ?{1pt} @{}}
    %\hline
    \Xhline{2\arrayrulewidth}
	\multirow{3}{0.8pt}{{\bf Filters}} & \multicolumn{2}{c ?{1pt}}{\bf DispNetC} & \multicolumn{2}{c?{1pt}}{ \bf PSMNet}  & \multicolumn{2}{c?{1pt}}{ \bf GANet} & \multicolumn{2}{c?{1pt}}{ \bf GCNet} \\
	\cline{2-9} % draw a horizontal line spanning only some of the table cells
	%&\multicolumn{8}{c}{Bad-3.0 Error Rates (\%)}  \\
	%\cline{2-9} % draw a horizontal line spanning only some of the table cells
	 &{noc}  &{all} &{noc}  &{all} &{noc}  &{all} &{noc}  &{all} \\
	%\hline \hline
	\Xhline{2\arrayrulewidth}
	W/O & 2.59 & 3.02 & 1.46  & 1.60  & \textbf{0.97}  & \textbf{1.10}   & 2.06  & 2.64  \\
	\Xhline{2\arrayrulewidth}
	\cellcolor{myskyblue} SABF & \cellcolor{mygray}2.26 & \cellcolor{mygray}2.63 & \cellcolor{mygray}1.28  & \cellcolor{mygray}1.40  & 1.07 & 1.17   & \cellcolor{mygray}1.76  & \cellcolor{mygray} 2.10  \\
	\cellcolor{myskyblue} DFN & \cellcolor{mygray} 2.37 & \cellcolor{mygray} 2.78  & \cellcolor{mygray} 1.23  & \cellcolor{mygray} 1.34  & 0.99  & 1.11   & \cellcolor{mygray} 1.70  & \cellcolor{mygray} 2.08  \\
	\cellcolor{myskyblue} PAC & \cellcolor{mygray} 2.38  & \cellcolor{mygray} 2.72  & \cellcolor{mygray} 1.29  & \cellcolor{mygray} 1.48  & 1.13  & 1.23   & \cellcolor{mygray} 1.71  & \cellcolor{mygray} 2.03  \\
	\cellcolor{myskyblue} SGA & \cellcolor{mygray}\textbf{1.90}  & \cellcolor{mygray}\textbf{2.18}  & \cellcolor{mygray}\textbf{1.17}  & \cellcolor{mygray}\textbf{1.32}  & - & -  & \cellcolor{mygray}\textbf{1.69}   & \cellcolor{mygray}\textbf{1.91}  \\
	\Xhline{2\arrayrulewidth}
	\end{tabularx}}
\vspace{-4pt}
\caption{KITTI 2015 bad-3 validation results. Improved results are highlighted in gray, and best ones are in bold. GANet contains SGA, resulting in blank entries ``-''.}
\label{tab:kt15-val30}
\end{table}

% move it back from the supplement
%\subsection{Runtime} 
\vspace{3pt} \noindent \textbf{Runtime} \, In Table \ref{tab:kt15-all-runtime}, we compare the GPU memory consumption and runtime in inference mode on pairs of frames with dimension $384 \times 1280$. All experiments are run on the same machine, with the same configuration of disparity range $D=192$, filter size $s=5$ and dilation rate $r=2$.

\vspace{3pt} \noindent \textbf{Comparison and Summary.} \, Our results show that most architectures benefit from adaptive filtering, with the exception of GANet, which already includes such filtering of SGA. It is worth pointing that GANet has the largest number of parameters and the longest runtime among the backbones. Lighter architectures, \eg PSMNet+SABF or PSMNet+DFN can achieve competitive results. DispNetC trails in terms of accuracy, but has about 20\% of the footprint of GANet and its combination with DFN strikes a good balance between accuracy and processing requirements.

% See methods effectiveness comparison on KITTI 2015 val-30 dataset in Table \ref{tab:model-effect-kt15-val30}.

\section{Conclusions}\label{sec:conclusions}
We demonstrate how deep adaptive or guided filtering can be integrated into representative 2D and 3D CNNs for stereo  matching with improved accuracy. %We integrate four filtering modules that are sensitive to pixel similarity or image edges and act like adaptive filters into four existing deep networks for stereo matching. 
Our extensive experimental results on Virtual KITTI 2 and KITTI 2015 highlight how our filtering modules effectively leverage the RGB information to dynamically guide the matching, resulting in further progress in stereo  matching. 
SGA, a component of GANet, is the most effective filtering mechanism and improves all backbones.
More broadly, our work shows that current state-of-the-art methods do not take full advantage of available information, with the exception of GANet which shows superior performance due to SGA and LGA, but has more parameters than GCNet and PSMNet. 
Integrating even the smaller filtering modules leads to decreases in error in the order of 10\%.

\iftrue
\begin{table}[b]
\small
\vspace{-8pt}
\centering
\scalebox{0.72}{
    %\begin{tabularx}{1.0\linewidth}{p{0.75cm} | XX XX XX XX}
    %\begin{tabularx}{1.0\linewidth}{@{} p{0.75cm} | YYY| YYY |YYY | YYY @{}}
    \begin{tabularx}{1.37\linewidth}{@{} ?{1pt} p{0.7cm} ?{1pt} YY ?{1pt} YY  ?{1pt}YY ?{1pt} YY ?{1pt} @{}}
    %\hline
    \Xhline{2\arrayrulewidth}
	\multirow{3}{0.8pt}{{\bf{Filters}}} & \multicolumn{2}{c ?{1pt} }{ \bf DispNetC} & \multicolumn{2}{c ?{1pt}  }{ \bf PSMNet}  & \multicolumn{2}{c ?{1pt}  }{ \bf GANet} & \multicolumn{2}{c ?{1pt}}{ \bf GCNet} \\
	\cline{2-9} % draw a horizontal line spanning only some of the table cells
	%&\multicolumn{8}{c}{Bad-3.0 Error Rates (\%)}  \\
	 &{Mem.}  &{Time}  &{Mem.}  &{Time} &{Mem.}  &{Time} &{Mem.}  &{Time} \\
	%\hline \hline
	\Xhline{2\arrayrulewidth}
	\hline
	W/O & \textbf{1394} & \textbf{18.35}  & \textbf{5151} &\textbf{315.57}  & \textbf{7178}& \textbf{1894.70}  & \textbf{4280} & \textbf{146.83}  \\
	\hline
	%\Xhline{2\arrayrulewidth}
	\cellcolor{myskyblue} SABF  & 1888 & 24.32  & 5386 & 563.42  & 7920 & 2488.72  & 4424 & 379.37   \\
	\cellcolor{myskyblue} DFN & 1422 & 28.33  & 5246 & 432.32  & 7466 & 2041.53  & 4298 & 255.20   \\
	\cellcolor{myskyblue}PAC & 1535 & 25.34  & 5168 & 514.91  & 8274 & 2383.44  & 4400 & 334.73   \\
	%\cellcolor{myskyblue}SGA & \cellcolor{mygray}\textbf{0.57} & \cellcolor{mygray}\textbf{9.37} & \cellcolor{mygray}\textbf{3.21} & \cellcolor{mygray}0.40 & \cellcolor{mygray}6.08 & \cellcolor{mygray}2.14  &- &- &-  & \cellcolor{mygray}\textbf{0.55} & \cellcolor{mygray}\textbf{9.24} & \cellcolor{mygray}\textbf{2.98}  \\
	\cellcolor{myskyblue} SGA & 7066 & 489.60  & 11070 & 823.00  & - & -  & 9916 & 655.18 \\
	%\hline
	\Xhline{2\arrayrulewidth}
	\end{tabularx}}
\vspace{-8pt}
\caption{Runtime (ms) and GPU memory consumption (MiB). Results are shown in the entries of filters (rows in skyblue) and backbones (columns) w.r.t. the baselines (rows W/O). The smallest values are in bold. GANet already contains SGA, resulting in blank entries ``-''.
}
\label{tab:kt15-all-runtime}
\end{table}
\fi

\noindent \textbf{Acknowledgements.} \, This research has been partially supported by National Science Foundation under Awards IIS-1527294 and IIS-1637761.

{\small
\bibliographystyle{ieee}
\bibliography{ref}
}

\newpage
\begin{center}
    \section*{\fontsize{13}{15}\selectfont Supplement}
    % regular font size is 12
    %\section*{\fontsize{12}{15}\selectfont Supplement}
    %\section*{Supplement}
\end{center}
In this supplementary material, we show the loss function, more details of filtering modules in the network, ablation study, and  additional qualitative results, some of which are mentioned but not fully discussed in the main paper due to the page limit.

%\subsection*{Loss Function and Network Details}
%\section{Loss Function}
\vspace{8pt} \noindent \textbf{Loss Function.} \, The loss is evaluated over the valid pixels which have ground truth disparity. We adopt the smooth $L_1$ loss function for end-to-end training:

\begin{equation}\label{eq:l1_loss_smooth}
\begin{aligned}[b]
L(d, \hat{d}) & = \frac{1}{N} \sum_{i=1}^N l_s(|| d_i - \hat{d_i} ||_1) \\
\text{and, } l_s(x) &=  \begin{cases} 0.5x^2 & \text{if } |x| < 1 \\ |x| - 0.5 & \text{otherwise}
\end{cases}
\end{aligned}
\end{equation}

\noindent where $N$ counts the valid pixels, $|| d_i - \hat{d_i} ||_1$ measures the absolute error of disparity prediction $\hat{d_i}$ and ground truth $d_i$.

\begin{figure}
\begin{center}
\includegraphics[width=1.0\linewidth]{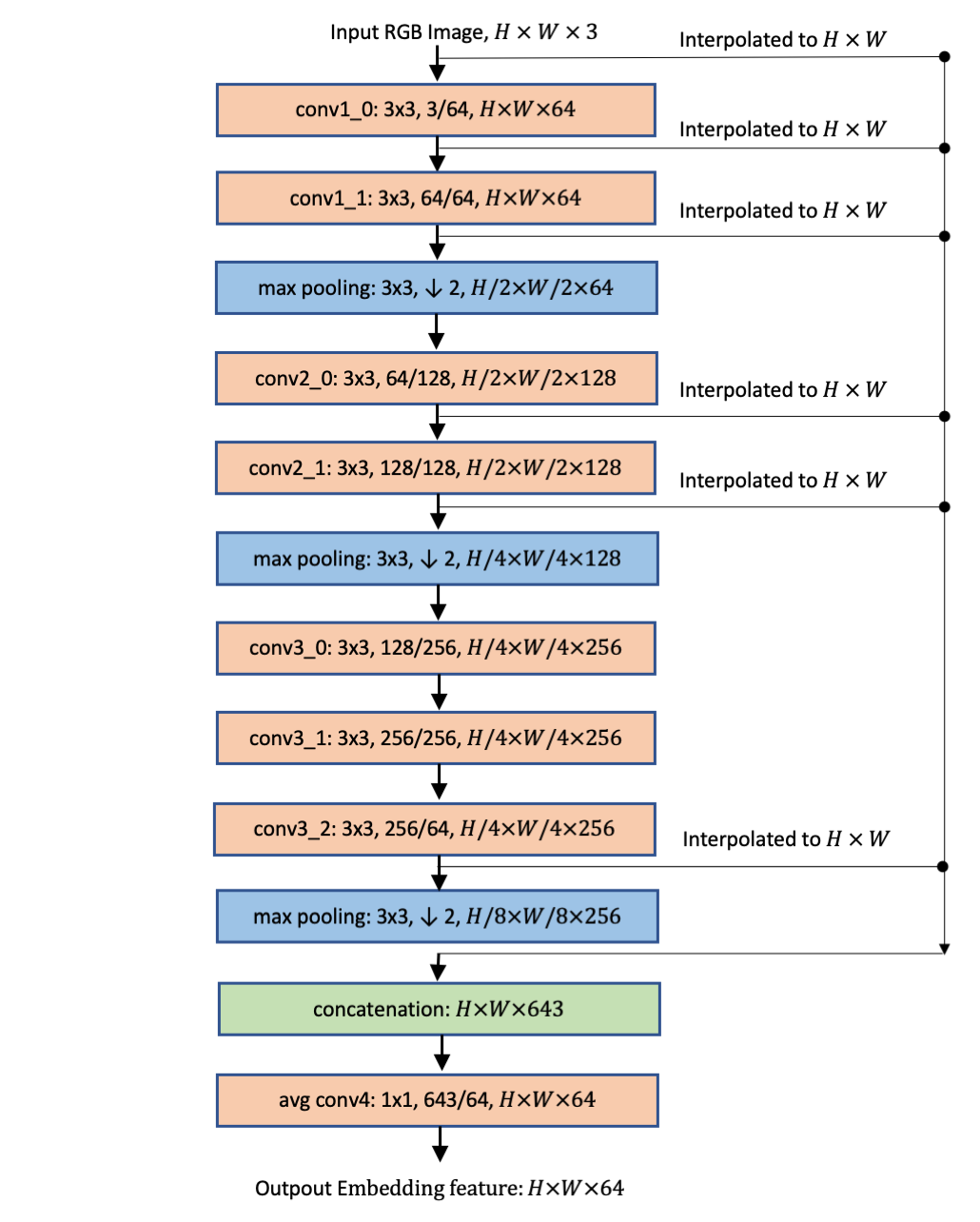}
\end{center}
\vspace{-10pt}
\caption{ 
The embedding network in the segmentation-aware bilateral filtering (SABF) module. Each convolutional layer is followed by a ReLU layer which is not drawn. Convolutional layers, \eg {\myqcrfont conv1\_0}, are defined by $3 \times 3$ as filter, $3/64$ as in/out feature planes, and $H \times W \times 64$ as the output dimension). Max pooling layers are implemented as $3 \times 3$ filter with stride of $2$ for downsampling (i.e., $\downarrow 2$). %Concatenation is performed along the feature channel.
}
\label{fig:sabf-archi}
\end{figure}

%\subsection*{Deep Adaptive Filtering Architectures} 
\vspace{6pt}\noindent \textbf{Deep Adaptive Filtering Architectures.} \, The integrated deep adaptive or guided filters include segmentation-aware bilateral filtering (SABF) \cite{harley_segaware2017}, dynamic filtering networks (DFN) \cite{jia2016dynamic}, pixel adaptive convolution (PAC) \cite{su2019pixel} and semi-global aggregation (SGA). Please see our source code (\url{https://github.com/ccj5351/DAFStereoNets}) for the detailed architectures. Fig. \ref{fig:sabf-archi} shows the embedding network of SABF.

%\subsection*{Ablation Study and Quantitative Results}
\vspace{6pt}\noindent \textbf{Ablation Study.} \,\, We perform ablation studies to investigate how the filter window $s$ and the dilation rate $r$ can affect the filtering output and disparity estimation. Out of a large number of possible combinations, we show two representatives \textit{DispNetC+SABF} and \textit{PSMNet+SABF} in Table \ref{tab:filter-size-kt15-val30}. We find that $s=5$ with $r=2$ achieve a good balance in accuracy, space and runtime. The 500-run averaged memory consumption and runtime in GPUs are measured when we test a $384 \times 1280$ stereo pair, and the \textit{bad-3} (noc,all) errors are evaluated on the KITTI 2015 validation set. Please note that a $5 \times 5$ filter (with dilation rate $2$) covers $9 \times 9$ regions in the cost feature space, which is equivalent to $33 \times 33$ regions in the RGB image space, due to the cost volume being a quarter of the size of the input images. In the following experiments, we keep using $s = 5$ with $r = 2$ for our different architectures. Please note that the results in Table \ref{tab:filter-size-kt15-val30} are obtained on a 16-core Intel Core i7-9800X CPU at 3.80GHz, and an NVIDIA TITAN Xp GPU with 12GB of RAM.

\vspace{6pt}\noindent \textbf{Effectiveness Comparison} \,\, In Table \ref{tab:ablation-effectiveness-kt15-val30}, we further investigate the effectiveness among those variants in terms of parameter increase (column $\delta P$\%) and error decrease (column $\delta E$\%) evaluated on the KITTI 2015 validation set, as shown in Table \ref{tab:supp-kt15-val30}. Please note that Table \ref{tab:supp-kt15-val30} is originally included in the main paper, and we repeat it here to explain the effectiveness in Table \ref{tab:ablation-effectiveness-kt15-val30}.

\begin{table*}
\small
\centering
\scalebox{0.97}{
    %\begin{tabularx}{1.0\linewidth}{p{1.0cm} | XX | XX | X | X | XX | XX | X | X}
    \begin{tabularx}{1.0\linewidth}{@{}  ?{1pt} p{1.cm} ?{1pt} YY ?{1pt} YY ?{1pt} Y ?{1pt} Y ?{1pt}  YY ?{1pt} YY ?{1pt} Y ?{1pt} Y ?{1pt} @{}}
    %\hline
    \Xhline{2\arrayrulewidth}
	\multirow{2}{1.0pt}{{\bf filter size $s$}} & \multicolumn{6}{c ?{1pt} }{\bf DispNetC+SABF} & \multicolumn{6}{c ?{1pt} }{\bf PSMNet+SABF} \\
	\cline{2-13} % draw a horizontal line spanning only some of the table cells
	& \multicolumn{2}{c ?{1pt} }{$r=1$} & \multicolumn{2}{c ?{1pt}} {$r=2$} & Mem. &  Time  & \multicolumn{2}{c ?{1pt}}{$r=1$} & \multicolumn{2}{c ?{1pt}} {$r=2$} & Mem. &  Time \\
\cline{2-13} % draw a horizontal line spanning only some of the table cells
	 & {EPE(px)} & {$\geq$3(\%)} & {EPE(px)} & {$\geq$3(\%)} & (MiB) & (ms) & {EPE(px)} & {$\geq$3(\%)} & {EPE(px)} & {$\geq$3(\%)} & (MiB) & (ms) \\
	%\hline
	\Xhline{2\arrayrulewidth}
	$s=3$ & 0.875 & 3.14 &  0.845 & 2.99  & 2132  & 45.26   & 0.639 & 1.63 &  0.643 & 1.61  & 4681  & 449.40  \\
	\hline
	%\rowcolor{gray} 
	%\cellcolor{gray}
	% \cellcolor[HTML]{AA0044}
	$s=5$ & 0.867 & 3.13 &   \cellcolor{mygray} 0.841  & \cellcolor{mygray} 2.90  & \cellcolor{mygray} 2228  & \cellcolor{mygray} 48.99 & 0.657 & 1.54 &   \cellcolor{mygray} 0.630  & \cellcolor{mygray} 1.46  & \cellcolor{mygray} 4989  & \cellcolor{mygray} 633.42
	\\
	\hline
	$s=7$ & 0.832 & 2.83 & 0.795  &  2.46 & 2588  & 54.21   & 0.650 & 1.50 & 0.642  &  1.54 & 4709  & 939.40  \\
	\hline
	$s=9$ & 0.825 & 2.84 &  0.854 & 3.00  & 3008  & 60.56  & 0.868 & 1.77 &  0.689 & 1.91  & 4953  & 1226.72\\
	%\hline 
	%$s=11$ & 0.843 & 2.91 &  0.826 & 3.03  & 3534  & 69.43   & - & - &  - & -  & -  & - \\
	%\hline
	\Xhline{2\arrayrulewidth}
	\end{tabularx}}
\vspace{-4pt}
\caption{Illustration of the effects of different filter window sizes $s$ and dilation rates $r$. We computer the bad-3 (noc,all) errors on the KITTI 2015 validation set and the averaged GPU memory consumption and runtime to test a  pair of frames with dimension $384 \times 1280$. %\PMcomments{Not very interesting and s=7 leads to better accuracy at small increase in time.} %\ccjscomments{ $s=11$ is not applicable to SABF+PSMNEt due to the GPU memory limit.}
}
\label{tab:filter-size-kt15-val30}
\end{table*}
%\begin{table}[b]
\begin{table}
\small
\centering
\scalebox{0.98}{
    %\begin{tabularx}{1.0\linewidth}{p{0.75cm} | XX XX XX XX}
    %\begin{tabularx}{1.0\linewidth}{@{} p{0.75cm} | YY| YY| YY| YY @{}}
    %note: ?{1pt} is used to draw a thick vertical line;
    \begin{tabularx}{1.0\linewidth}{@{} ?{1pt} p{0.75cm}?{1pt}  YY ?{1pt}  YY ?{1pt} YY ?{1pt} YY ?{1pt} @{}}
    %\hline
    \Xhline{2\arrayrulewidth}
	\multirow{3}{0.8pt}{{\bf Filters}} & \multicolumn{2}{c ?{1pt}}{\bf DispNetC} & \multicolumn{2}{c?{1pt}}{ \bf PSMNet}  & \multicolumn{2}{c?{1pt}}{ \bf GANet} & \multicolumn{2}{c?{1pt}}{ \bf GCNet} \\
	\cline{2-9} % draw a horizontal line spanning only some of the table cells
	%&\multicolumn{8}{c}{Bad-3.0 Error Rates (\%)}  \\
	%\cline{2-9} % draw a horizontal line spanning only some of the table cells
	 &{noc}  &{all} &{noc}  &{all} &{noc}  &{all} &{noc}  &{all} \\
	%\hline \hline
	\Xhline{2\arrayrulewidth}
	W/O & 2.59 & 3.02 & 1.46  & 1.60  & \textbf{0.97}  & \textbf{1.10}   & 2.06  & 2.64  \\
	\Xhline{2\arrayrulewidth}
	\cellcolor{myskyblue} SABF & \cellcolor{mygray}2.26 & \cellcolor{mygray}2.63 & \cellcolor{mygray}1.28  & \cellcolor{mygray}1.40  & 1.07 & 1.17   & \cellcolor{mygray}1.76  & \cellcolor{mygray} 2.10  \\
	\cellcolor{myskyblue} DFN & \cellcolor{mygray} 2.37 & \cellcolor{mygray} 2.78  & \cellcolor{mygray} 1.23  & \cellcolor{mygray} 1.34  & 0.99  & 1.11   & \cellcolor{mygray} 1.70  & \cellcolor{mygray} 2.08  \\
	\cellcolor{myskyblue} PAC & \cellcolor{mygray} 2.38  & \cellcolor{mygray} 2.72  & \cellcolor{mygray} 1.29  & \cellcolor{mygray} 1.48  & 1.13  & 1.23   & \cellcolor{mygray} 1.71  & \cellcolor{mygray} 2.03  \\
	\cellcolor{myskyblue} SGA & \cellcolor{mygray}\textbf{1.90}  & \cellcolor{mygray}\textbf{2.18}  & \cellcolor{mygray}\textbf{1.17}  & \cellcolor{mygray}\textbf{1.32}  & - & -  & \cellcolor{mygray}\textbf{1.69}   & \cellcolor{mygray}\textbf{1.91}  \\
	\Xhline{2\arrayrulewidth}
	\end{tabularx}}
\vspace{-4pt}
\caption{KITTI 2015 bad-3 validation results. Improved results are highlighted in gray, and best ones are in bold. GANet contains SGA, resulting in blank entries ``-''.}
\label{tab:supp-kt15-val30}
\end{table}

\begin{table}
\small
\centering
\scalebox{0.86}{

   %\begin{tabularx}{1.0\linewidth}{@{} ?{1pt} p{0.75cm}?{1pt}  YY ?{1pt}  YY ?{1pt} YY ?{1pt} YY ?{1pt} @{}}
   \begin{tabularx}{1.15\linewidth}{@{} ?{1pt} p{0.7cm}?{1pt}  YY |  YY | YY | YY ?{1pt} @{}}
    %\begin{tabularx}{\linewidth}{ @{} ?{1pt} p{0.8cm} | YY YY YY YY  ?{1pt} @{} }
    %\hline
    \Xhline{2\arrayrulewidth}
	\multirow{2}{0.6pt}{{\bf Filters}} & \multicolumn{2}{c ?{1pt}}{\bf DispNetC} & \multicolumn{2}{c?{1pt}}{ \bf PSMNet}  & \multicolumn{2}{c?{1pt}}{ \bf GANet} & \multicolumn{2}{c?{1pt}}{ \bf GCNet} \\
	\cline{2-9} % draw a horizontal line spanning only some of the table cells
	%&\multicolumn{8}{c}{Bad-3.0 Error Rates (\%)}  \\
	%\cline{2-9} % draw a horizontal line spanning only some of the table cells
	&{$\delta$E(\%)} & {$\delta$P(\%)} &{$\delta$E(\%)} & {$\delta$P(\%)} &{$\delta$E(\%)} & {$\delta$P(\%)} &{$\delta$E(\%)} & {$\delta$P(\%)} \\
	%\hline \hline
	%\Xhline{2\arrayrulewidth}
	%W/O & - & - &  -  & - & - & -  & - & -  \\
	\Xhline{2\arrayrulewidth}
	SABF & 12.9 & 4.2 & 12.4 & 34 &  {-5.9} & 27  & 20.6 & 62.4 \\
	DFN & 7.9 & 0.8 & 16.2 & 6.4 & -0.1 & 5.1  & 21.5 & 11.8 \\
	PAC & 9.9 & 0.1 & 7.8 & 2.0 & -12 & 1.6  & 23.3 & 3.6 \\
	SGA & 27.8 & 7.0 & 17.7 & 58.8 & - & - & 27.7 & 108 \\
	%\hline 
	\Xhline{2\arrayrulewidth}
	\end{tabularx}
	}
\vspace{-4pt}

\caption{Effectiveness comparison on the  KITTI 2015 val-30 dataset. For each combination of network backbone and filtering, columns $\delta$\textit{E}(\%) and $\delta$\textit{P}(\%) indicate the relative decrease of error and increase of the number of parameters, respectively, w.r.t. the backbone baselines.}
\label{tab:ablation-effectiveness-kt15-val30}
\end{table}

\begin{figure*}[t]
    \centering
    \renewcommand{\tabcolsep}{0.2pt}
    \scriptsize
    \begin{tabular}{ccc}
        \includegraphics[width=0.33\textwidth]{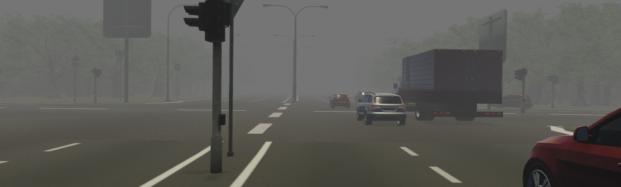} &
        
        \begin{overpic}[width=0.33\textwidth]{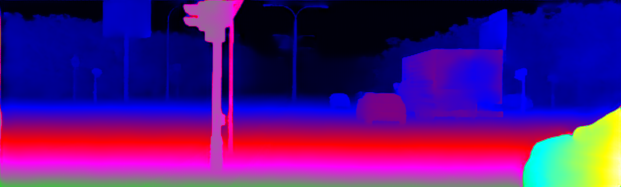}
        \put (85,25) {$\displaystyle\textcolor{white}{\textbf{7.19\%}}$}
        \put(75,17){{\color{white} \circle{18}}}
        \end{overpic} & 
        
        \begin{overpic}[width=0.33\textwidth]{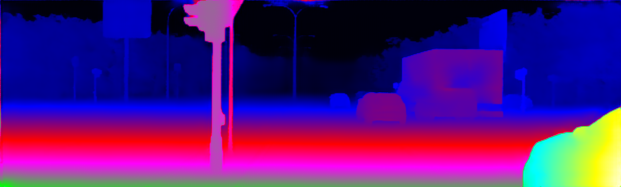}
        \put (85,25) {$\displaystyle\textcolor{white}{\textbf{4.66\%}}$}
        \put(75,17){{\color{white} \circle{18}}}
        \end{overpic}
        
        \\
        \includegraphics[width=0.33\textwidth]{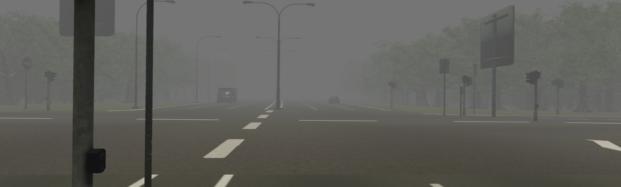} &
         \begin{overpic}[width=0.33\textwidth]{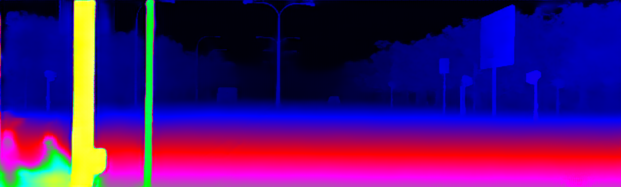}
        \put (85,25) {$\displaystyle\textcolor{white}{\textbf{3.03\%}}$}
        \put(6,6){{\color{white} \circle{12}}}
        \end{overpic} & 
        
        \begin{overpic}[width=0.33\textwidth]{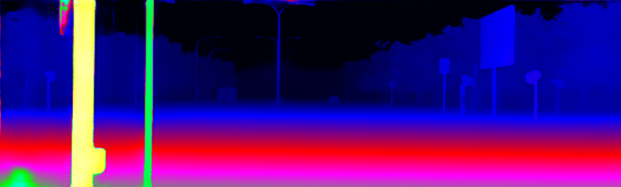}
        \put (85,25) {$\displaystyle\textcolor{white}{\textbf{2.43\%}}$}
        \put(6,6){{\color{white} \circle{12}}}
        \end{overpic}
        
        \\
        & \normalsize (a) & \\ 
        \includegraphics[width=0.33\textwidth]{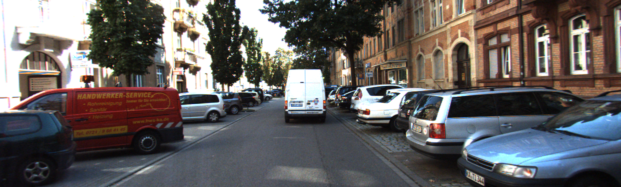} &
        
        \begin{overpic}[width=0.33\textwidth]{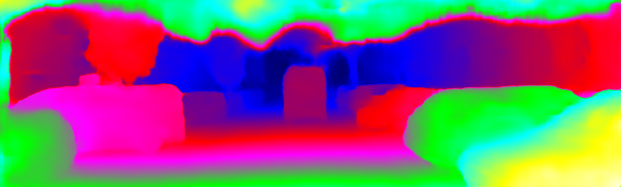}
        \put (85,23) {$\displaystyle\textcolor{white}{\textbf{7.59\%}}$}
        \put(13,17){{\color{white} \circle{12}}}
        \end{overpic} & 
        
        \begin{overpic}[width=0.33\textwidth]{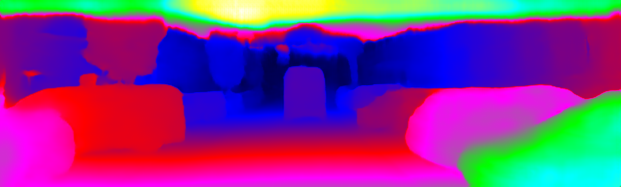}
        \put (85,23) {$\displaystyle\textcolor{white}{\textbf{5.42\%}}$}
        \put(13,17){{\color{white} \circle{12}}}
        \end{overpic} \\
        
        \includegraphics[width=0.33\textwidth]{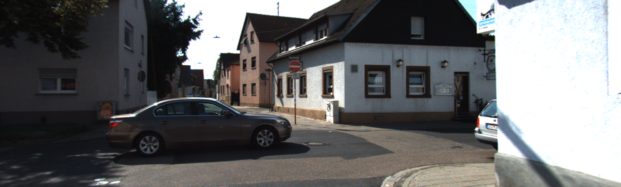} &
        
        \begin{overpic}[width=0.33\textwidth]{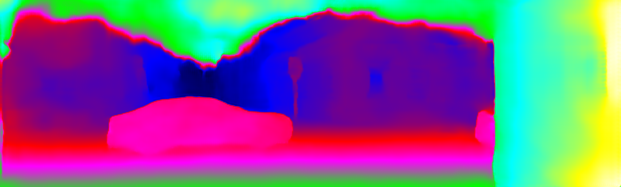}
        \put (5,20) {$\displaystyle\textcolor{white}{\textbf{9.23\%}}$}
        \put(88,15){{\color{white} \circle{22}}}
        \end{overpic} & 
        
        \begin{overpic}[width=0.33\textwidth]{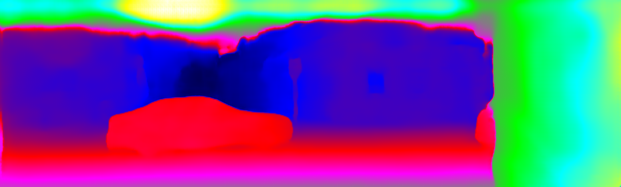}
        \put (5,20) {$\displaystyle\textcolor{white}{\textbf{7.35\%}}$}
        \put(88,15){{\color{white} \circle{22}}}
        \end{overpic}
        \\
        & \normalsize (b) &
    \end{tabular}
    \vspace{-8pt}
    \caption{Results using DispNetC as backbone. (a) DispNetC vs DispNetC+PAC on Virtual KITTI 2 Scene06 validation set. (b) DispNetC vs DispNetC+SABF on KITTI 2015 validation set. In all rows, the left image is the reference image of the stereo pair, the middle column in the disparity map from the unmodified backbone, and the right image is the disparity map of the backbone with the integrated filter.}
    \label{fig:disparitymaps-backbone-dispnetc}
\end{figure*}

\begin{figure*}[t]
    \centering
    \renewcommand{\tabcolsep}{0.2pt}
    \scriptsize
    \begin{tabular}{ccc}
         \includegraphics[width=0.33\textwidth]{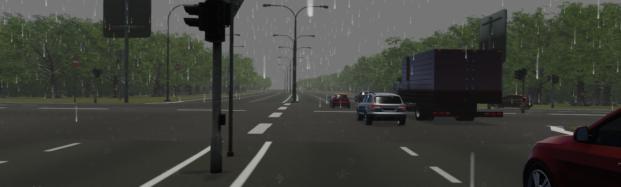} &
        
        \begin{overpic}[width=0.33\textwidth]{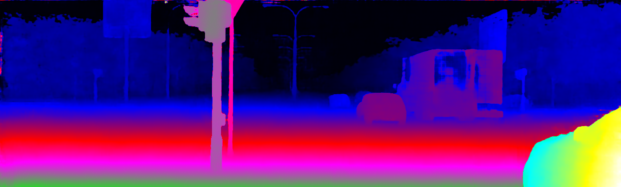}
        \put (4,25) {$\displaystyle\textcolor{white}{\textbf{3.84\%}}$}
        \put(75,17){{\color{white} \circle{18}}}
        \end{overpic} & 
        
        \begin{overpic}[width=0.33\textwidth]{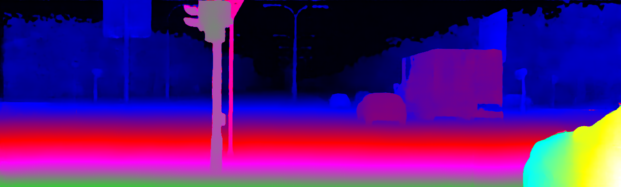}
        \put (4,25) {$\displaystyle\textcolor{white}{\textbf{2.03\%}}$}
        \put(75,17){{\color{white} \circle{18}}}
        \end{overpic}
        
        \\
        \includegraphics[width=0.33\textwidth]{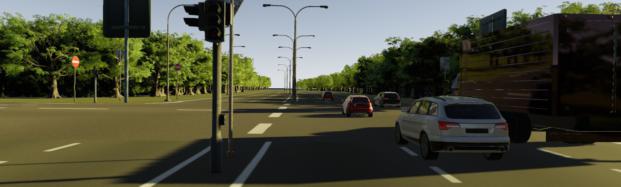} &
        
        \begin{overpic}[width=0.33\textwidth]{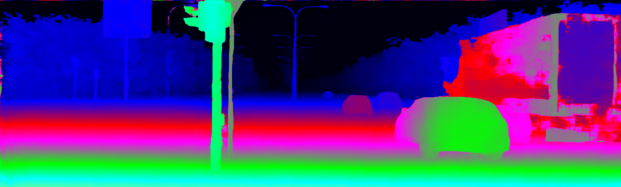}
        \put (4,25) {$\displaystyle\textcolor{white}{\textbf{11.50\%}}$}
        \put(90,19){{\color{white} \circle{18}}}
        \end{overpic} & 
        
        \begin{overpic}[width=0.33\textwidth]{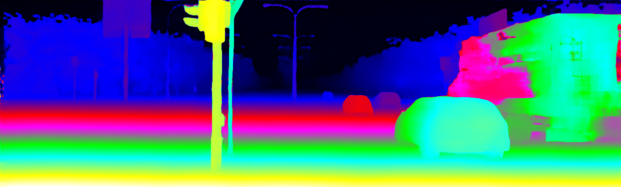}
        \put (4,25) {$\displaystyle\textcolor{white}{\textbf{6.69\%}}$}
        \put(90,19){{\color{white} \circle{18}}}
        \end{overpic}
        \\
        
        & \normalsize (a) & \\ 
        
        \includegraphics[width=0.33\textwidth]{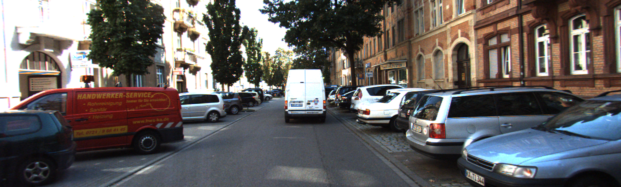} &
        
        \begin{overpic}[width=0.33\textwidth]{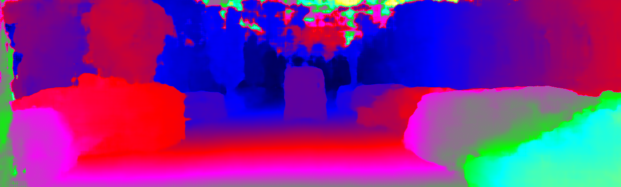}
        \put (4,25) {$\displaystyle\textcolor{white}{\textbf{6.94\%}}$}
        \put(53,24){{\color{white} \circle{15}}}
        \end{overpic} & 
        
        \begin{overpic}[width=0.33\textwidth]{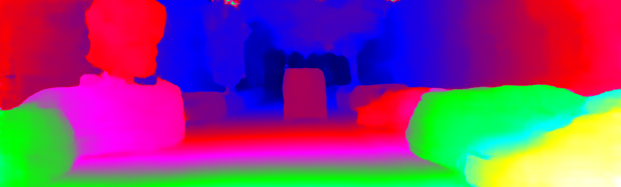}
        \put (4,25) {$\displaystyle\textcolor{white}{\textbf{4.82\%}}$}
        \put(53,24){{\color{white} \circle{15}}}
        \end{overpic} \\
         
         \includegraphics[width=0.33\textwidth]{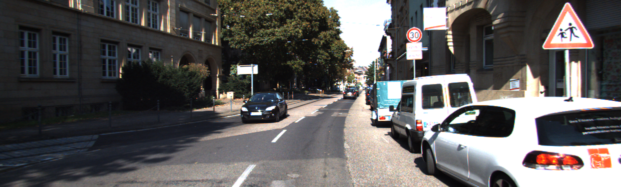} &
        
        \begin{overpic}[width=0.33\textwidth]{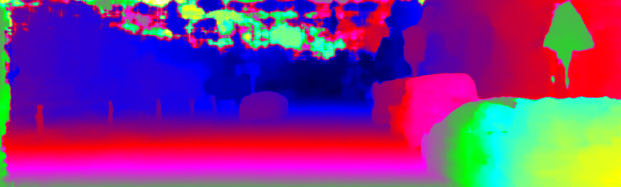}
        \put (4,25) {$\displaystyle\textcolor{white}{\textbf{4.03\%}}$}
        \put(50,24){{\color{white} \circle{15}}}
        \end{overpic} & 
        
        \begin{overpic}[width=0.33\textwidth]{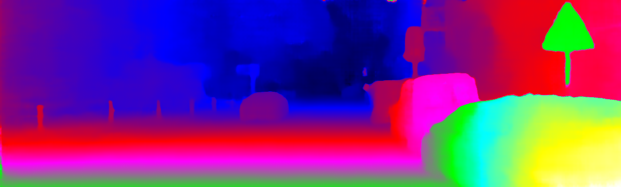}
        \put (4,25) {$\displaystyle\textcolor{white}{\textbf{2.54\%}}$}
        \put(50,24){{\color{white} \circle{15}}}
        \end{overpic} 
        
        \\
        & \normalsize (b) &
    \end{tabular}
    \vspace{-8pt}
    \caption{Results using  GCNet as backbone. (a) GCNet vs GCNet+SGA on Virtual KITTI 2 Scene06 validation set. (b) GCNet vs GCNet+SGA on KITTI 2015 validation set. In all rows, the left image is the reference image of the stereo pair, the middle column in the disparity map from the unmodified backbone, and the right image is the disparity map of the backbone with the integrated filter.}
    % NOTE: both from SGA-GCNet vs GCNet!!!
    \label{fig:disparitymaps-backbone-gcnet}
\end{figure*}

\begin{figure*}[t]
    \centering
    \renewcommand{\tabcolsep}{0.2pt}
    \scriptsize
    \begin{tabular}{ccc}
        \includegraphics[width=0.33\textwidth]{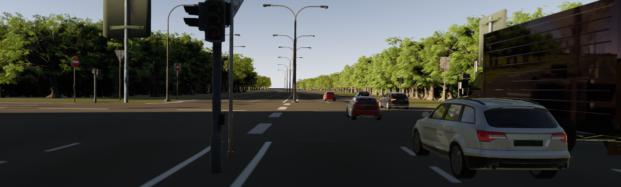} &
        
        \begin{overpic}[width=0.33\textwidth]{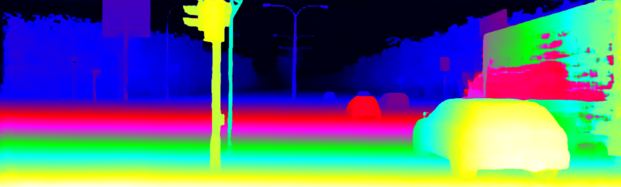}
        \put (2,25) {$\displaystyle\textcolor{white}{\textbf{4.71\%}}$}
        \put(87,17){{\color{white} \circle{18}}}
        \end{overpic} & 
        
        \begin{overpic}[width=0.33\textwidth]{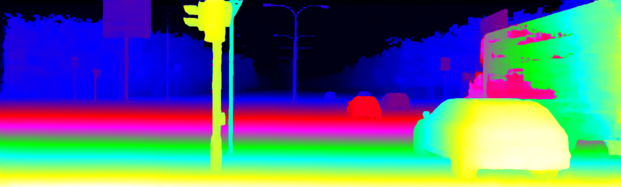}
        \put (2,25) {$\displaystyle\textcolor{white}{\textbf{2.70\%}}$}
        \put(87,17){{\color{white} \circle{18}}}
        \end{overpic} 
        \\
         \includegraphics[width=0.33\textwidth]{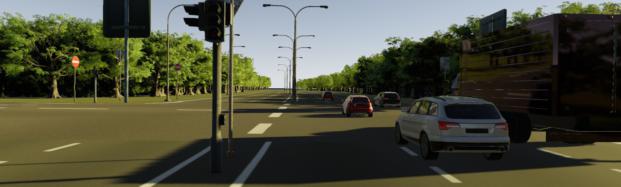} &
        
        \begin{overpic}[width=0.33\textwidth]{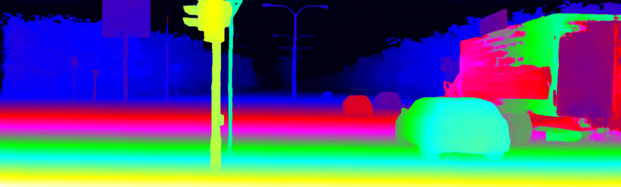}
        \put (2,25) {$\displaystyle\textcolor{white}{\textbf{12.23\%}}$}
        \put(90,17){{\color{white} \circle{18}}}
        \end{overpic} & 
        
        \begin{overpic}[width=0.33\textwidth]{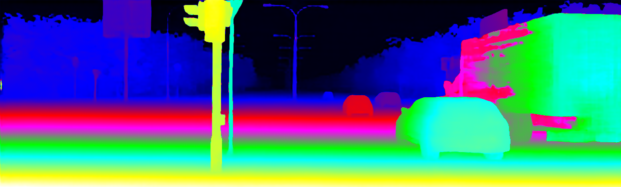}
        \put (2,25) {$\displaystyle\textcolor{white}{\textbf{3.81\%}}$}
        \put(90,17){{\color{white} \circle{18}}}
        \end{overpic} 
        
        \\
        & \normalsize (a) & \\ 
        
         \includegraphics[width=0.33\textwidth]{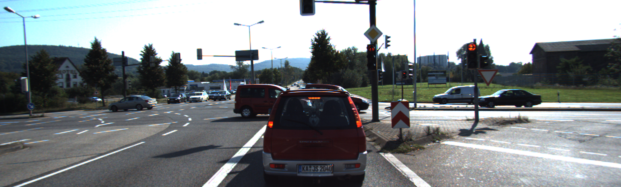} &
        
        \begin{overpic}[width=0.33\textwidth]{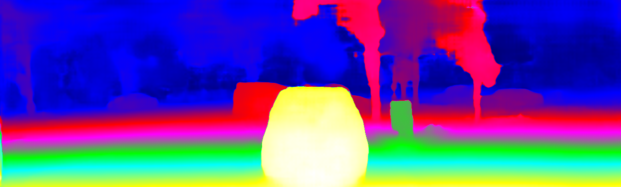}
        \put (2,25) {$\displaystyle\textcolor{white}{\textbf{2.29\%}}$}
        \put(50,16){{\color{white} \circle{10}}}
        \put(55,8){{\color{white} \circle{12}}}
        \end{overpic} & 
        
        \begin{overpic}[width=0.33\textwidth]{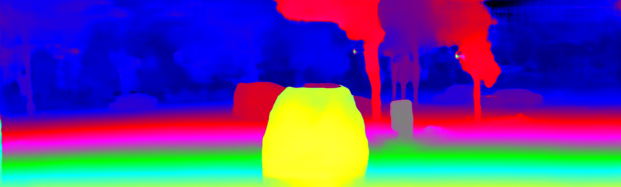}
        \put (2,25) {$\displaystyle\textcolor{white}{\textbf{1.21\%}}$}
        \put(50,16){{\color{white} \circle{10}}}
        \put(55,8){{\color{white} \circle{12}}}
        \end{overpic} 
        
        \\ 
        \includegraphics[width=0.33\textwidth]{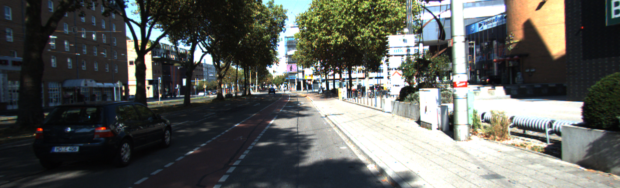} &
        
        \begin{overpic}[width=0.33\textwidth]{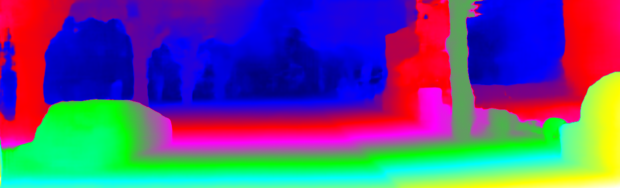}
        \put (2,25) {$\displaystyle\textcolor{white}{\textbf{3.02\%}}$}
        \put(8,8){{\color{white} \circle{14}}}
        \end{overpic} & 
        
        \begin{overpic}[width=0.33\textwidth]{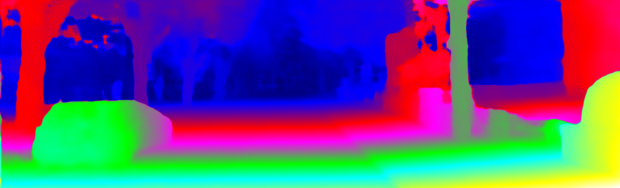}
        \put (2,25) {$\displaystyle\textcolor{white}{\textbf{1.71\%}}$}
        \put(8,8){{\color{white} \circle{14}}}
        \end{overpic} 
        \\
        & \normalsize (b) &
    \end{tabular}
    \vspace{-8pt}
    \caption{Results using PSMNet as backbone.  (a) PSMNet vs PSMNet+DFN on Virtual KITTI 2 Scene06 validation set.(b) PSMNet vs PSMNet+PAC on KITTI 2015 validation set. In all rows, the left image is the reference image of the stereo pair, the middle column in the disparity map from the unmodified backbone, and the right image is the disparity map of the backbone with the integrated filter.}
    \label{fig:disparitymaps-backbone-psmnet}
\end{figure*}

\begin{figure*}[t]
    \centering
    \renewcommand{\tabcolsep}{0.2pt}
    \scriptsize
    \begin{tabular}{ccc}
        %\includegraphics[width=0.33\textwidth]{img/ganet-img-0050.jpg} &
        %\begin{overpic}[width=0.33\textwidth]{img/ganet-disp-color-0050.png}
        %\put (2,25) {$\displaystyle\textcolor{white}{\textbf{0.49\%}}$}
        %\put(40,17){{\color{white} \circle{10}}}
        %\end{overpic} & 
        
        %\begin{overpic}[width=0.33\textwidth]{img/sabf-ganet-disp-color-0050.png}
        %\put (2,25) {$\displaystyle\textcolor{white}{\textbf{0.51\%}}$}
        %\put(40,17){{\color{white} \circle{10}}}
        %\end{overpic}
        
        %\\
        
        %\includegraphics[width=0.33\textwidth]{img/ganet-img-2660.jpg} &
        %\begin{overpic}[width=0.33\textwidth]{img/ganet-disp-color-2660.png}
        %\put (2,25) {$\displaystyle\textcolor{white}{\textbf{0.51\%}}$}
        %\put(43,25){{\color{white} \circle{10}}}
        %\end{overpic} & 
        
        %\begin{overpic}[width=0.33\textwidth]{img/sabf-ganet-disp-color-2660.png}
        %\put (2,25) {$\displaystyle\textcolor{white}{\textbf{0.51\%}}$}
        %\put(43,25){{\color{white} \circle{10}}}
        %\end{overpic} \\

        \includegraphics[width=0.33\textwidth]{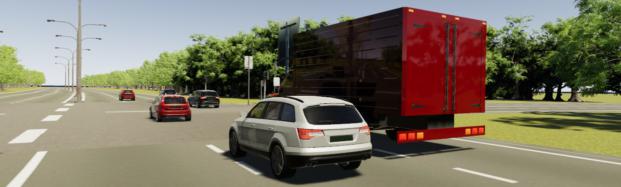} &
        \begin{overpic}[width=0.33\textwidth]{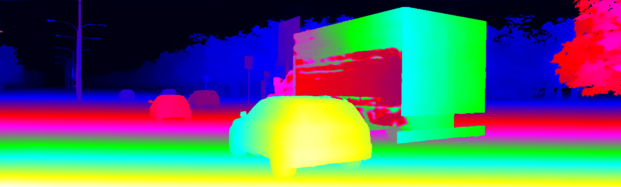}
        \put (2,25) {$\displaystyle\textcolor{white}{\textbf{6.34\%}}$}
        \put(54,15){{\color{white} \circle{20}}}
        \end{overpic} & 
        
        \begin{overpic}[width=0.33\textwidth]{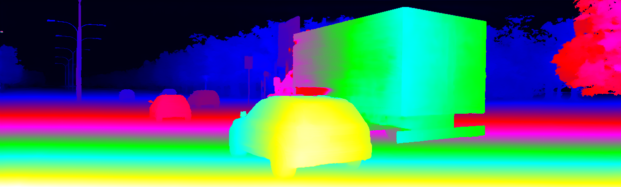}
        \put (2,25) {$\displaystyle\textcolor{white}{\textbf{2.33\%}}$}
        \put(54,15){{\color{white} \circle{20}}}
        \end{overpic}
        
        \\
        
        \includegraphics[width=0.33\textwidth]{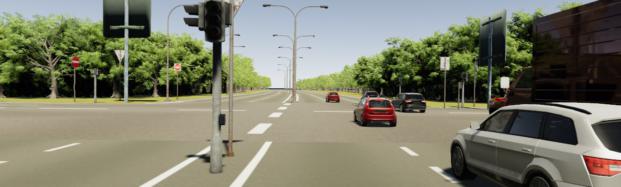} &
        
        \begin{overpic}[width=0.33\textwidth]{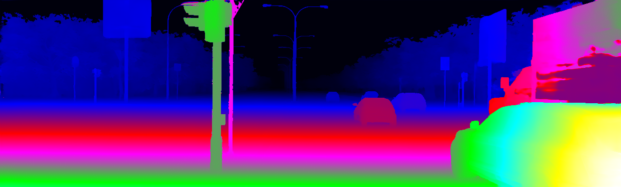}
        \put (2,25) {$\displaystyle\textcolor{white}{\textbf{4.30\%}}$}
        \put(93,15){{\color{white} \circle{12}}}
        \end{overpic} & 
        
        \begin{overpic}[width=0.33\textwidth]{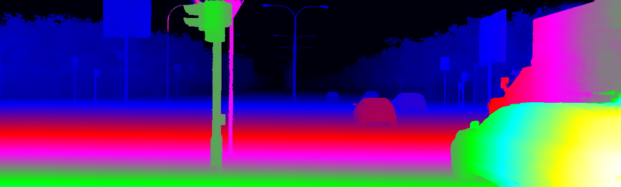}
        \put (2,25) {$\displaystyle\textcolor{white}{\textbf{1.22\%}}$}
        \put(93,15){{\color{white} \circle{12}}}
        \end{overpic}
        
        \\
        & \normalsize (a) & \\ 
       
       \includegraphics[width=0.33\textwidth]{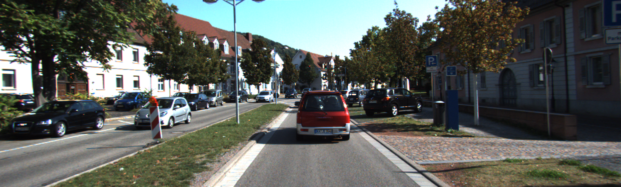} &
        
        \begin{overpic}[width=0.33\textwidth]{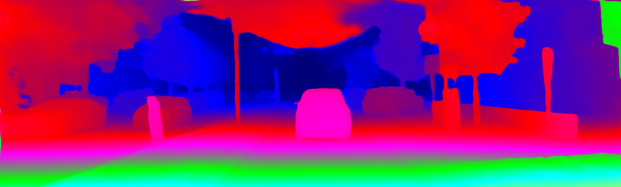}
        \put (2,25) {$\displaystyle\textcolor{white}{\textbf{1.79\%}}$}
        \put(7,16){{\color{white} \circle{12}}}
        \end{overpic} & 
        
        \begin{overpic}[width=0.33\textwidth]{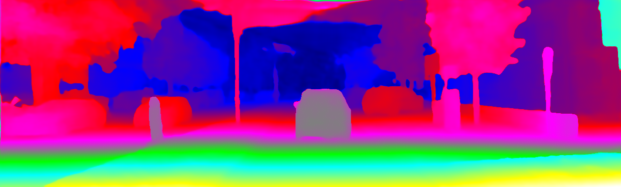}
        \put (2,25) {$\displaystyle\textcolor{white}{\textbf{1.12\%}}$}
        \put(7,16){{\color{white} \circle{12}}}
        \end{overpic} \\
        
         \includegraphics[width=0.33\textwidth]{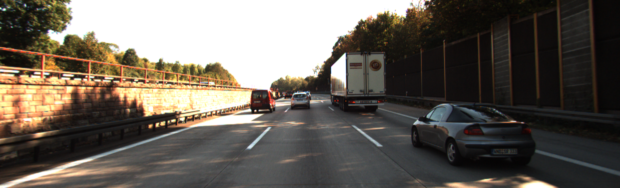} &
        \begin{overpic}[width=0.33\textwidth]{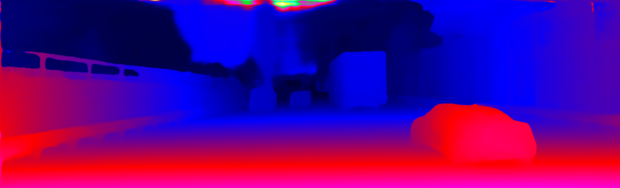}
        \put (10,25) {$\displaystyle\textcolor{white}{\textbf{1.07\%}}$}
        \put(5,20){{\color{white} \circle{10}}}
        \end{overpic} & 
        
        \begin{overpic}[width=0.33\textwidth]{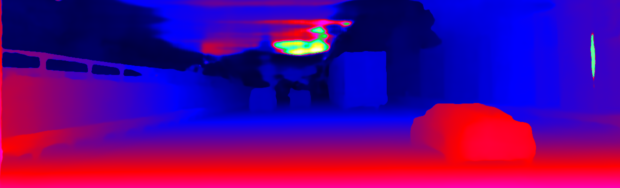}
        \put (10,25) {$\displaystyle\textcolor{white}{\textbf{0.95\%}}$}
        \put(5,20){{\color{white} \circle{10}}}
        \end{overpic}

        \\
        & \normalsize (b) &
    \end{tabular}
    \vspace{-8pt}
    \caption{Results using  GANet as backbone. (a) GANet vs GANet+SABF on Virtual KITTI 2 Scene06 validation set. (b) GANet vs GANet+PAC on KITTI 2015 validation set. In all rows, the left image is the reference image of the stereo pair, the middle column in the disparity map from the unmodified backbone, and the right image is the disparity map of the backbone with the integrated filter.}
    \label{fig:disparitymaps-backbone-ganet}
\end{figure*}

%\subsection*{Qualitative Results}
\vspace{6pt}\noindent \textbf{Qualitative Results} \,\, In Figs. \ref{fig:disparitymaps-backbone-dispnetc}--\ref{fig:disparitymaps-backbone-ganet}, we show reference images and disparity maps generated by each backbone without modifications and the same backbone after integrating one of the filtering techniques.

\end{document}